\documentclass[sigconf,natbib=true,anonymous=false]{acmart}
\acmConference[XX '26]{XXX}{XX XX--XX, XXXX}{XXX}
\usepackage[utf8]{inputenc} % allow utf-8 input
\usepackage[T1]{fontenc}    % use 8-bit T1 fonts
\usepackage{hyperref}       % hyperlinks
\usepackage{url}            % simple URL typesetting
\usepackage{booktabs}       % professional-quality tables
\usepackage{amsthm}
\usepackage{amsfonts}       % blackboard math symbols
\usepackage{nicefrac}       % compact symbols for 1/2, etc.
\usepackage{microtype}      % microtypography

\usepackage{amsmath, amssymb}
\usepackage{bbding}
\usepackage{tikz}
\usetikzlibrary{backgrounds,arrows.meta,shapes,shadows}
\usepackage{pgfplots}
\usepackage{tabularx}
\pgfplotsset{compat=newest,compat/show suggested version=false}

\usepackage{cleveref}
\usepackage{multirow}
\usepackage{fontawesome}
\usepackage[normalem]{ulem}
\usepackage{makecell}
\usepackage{xurl}

\usepackage{subcaption} % for subfigure

\usepackage{array}
\usepackage{pifont}

\definecolor{c01-blue}{HTML}{1F77B4}
\definecolor{c02-orange}{HTML}{FF7F0E}
\definecolor{c03-green}{HTML}{2CA02C}
\definecolor{c04-red}{HTML}{D62728}
\definecolor{c05-purple}{HTML}{9467BD}
\definecolor{c06-brown}{HTML}{8C564B}
\definecolor{c07-pink}{HTML}{E377C2}
\definecolor{c08-gray}{HTML}{7F7F7F}
\definecolor{c09-yellow}{HTML}{BCBD22}
\definecolor{c10-cyan}{HTML}{17BECF}
\newenvironment{qaxis}[1][]%
  {\begin{axis}[
    scale only axis,
    grid=major,
    grid style={gray,opacity=0.2},
    ylabel={Throughput (tokens/s)},
    xtick=data,
    line width=.75pt,
    xtick={1,2,3,4,5,6},
    xticklabels={\q{q2},\q{q3},\q{q4},\q{q5},\q{q8},\q{fp16}},
    every major tick/.append style={major tick length=2pt,black},
    every minor tick/.append style={minor tick length=1.5pt,gray},
    xlabel style={},
    x tick label style={
      /pgf/number format/assume math mode,
      /pgf/number format/1000 sep={}},
    y tick label style={
      /pgf/number format/assume math mode,
      /pgf/number format/1000 sep={}},
    xlabel style={yshift=0pt},
    ylabel style={yshift=-4pt},
    legend style={at={(rel axis cs:.5,1.05)},
      draw=none,align=left,
      anchor=south,fill=none,
      legend columns=-1},
    mark options={mark size=.8pt,solid},
    #1
  ]}%
  {\end{axis}}%

  {\begin{qaxis}[
    xtick={1,2,3,4,5,6},
    xticklabels={64,128,256,512},
    #1
  ]}%
  {\end{qaxis}}%

  {\end{axis}}%

\newcommand{\q}[1]{\texttt{\MakeLowercase{#1}}}

\begin{document}

\title{A Systematic Evaluation of On-Device LLMs: Quantization, Performance, and Resources}

\author{%
\small
Qingyu Song$^{1}$,
Rui Liu$^{2}$,
Wei Lin$^{3}$,
Peiyu Liao$^{4}$,
Wenqian Zhao$^{4}$,
Yiwen Wang$^{4}$,
Shoubo Hu$^{4}$,
Yining Jiang$^{1}$,\\
Mochun Long$^{1}$,
Hui-Ling Zhen$^{4}$,
Ning Jiang$^{1}$,
Mingxuan Yuan$^{4}$,
Qiao Xiang$^{1*}$,
Hong Xu$^{3*}$
}

\thanks{*Corresponding authors: qiaoxiang@xmu.edu.cn, hongxu@cuhk.edu.hk}

\affiliation{%
  \institution{%
  \small
    $^{1}$Xiamen University,
    $^{2}$Guangdong University of Technology,
    $^{3}$The Chinese University of Hong Kong,
    $^{4}$Huawei
  }
  \country{}
}

\renewcommand{\shortauthors}{Song et al.}
\settopmatter{printacmref=false}
\renewcommand\footnotetextcopyrightpermission[1]{}


\begin{abstract}
  Deploying Large Language Models (LLMs) on edge devices enhances privacy but faces performance hurdles due to limited resources. We introduce a systematic methodology to evaluate on-device LLMs, balancing capability, efficiency, and resource constraints. Through an extensive analysis of models (0.5B–14B) and seven post-training quantization (PTQ) methods on commodity hardware, we demonstrate that: 1) Heavily quantized large models consistently outperform smaller, high-precision models, with a performance threshold at $\sim$3.5 effective bits-per-weight (BPW); 2) Resource utilization scales linearly with BPW, though power and memory footprints vary by quantization algorithm; and 3) With a reduction in model size, the primary constraint on throughput transitions from communication overhead to computational latency. We conclude by offering guidelines for optimizing LLMs in resource-constrained edge environments. Our codebase is available at  \url{https://anonymous.4open.science/r/LLMOnDevice/}.
\end{abstract}

\begin{CCSXML}
<ccs2012>
   <concept>
       <concept_id>10002944.10011123.10011130</concept_id>
       <concept_desc>General and reference~Evaluation</concept_desc>
       <concept_significance>500</concept_significance>
       </concept>
 </ccs2012>
\end{CCSXML}

\ccsdesc[500]{General and reference~Evaluation}

\keywords{On-Device LLM, Inference, Evaluation}


\maketitle

\section{Introduction} \label{sec:introduction}

LLMs have revolutionized modern applications through their advanced text-generation capabilities~\citep{achiam2023gpt, liu2024deepseek}.
While traditionally cloud-dependent, a paradigm shift towards on-device deployment is emerging~\citep{chen2024octopus, hu2024minicpm}, enabled by breakthroughs in efficient model design~\citep{llm_compression_survey} and advancements in edge computing hardware.
% While cloud-based deployment has historically dominated due to computational demands, recent trends reveal a paradigm shift toward on-device LLM. 
% This transition is driven by two synergistic factors: the continuous enhancement of smaller-scale models through architectural innovations and the steady improvement of edge-side computing capabilities in consumer hardware. 
This transition also addresses critical privacy concerns: on-device execution eliminates remote data transmission, positioning these models as essential tools for privacy-sensitive domains such as healthcare~\citep{thirunavukarasu2023large} and finance~\citep{wu2023bloomberggpt}. 
% Crucially, on-device LLMs offer a compelling privacy-preserving alternative to cloud-dependent systems, as sensitive user data no longer needs to be transmitted to remote servers. 
% This inherent advantage positions on-device models as a critical enabler for privacy-sensitive applications in healthcare, finance, and personal productivity domains.
% \qy{citation}

% motivation
Despite these advantages, on-device LLMs face inherent limitations due to their reduced parameter count and the use of compression techniques like quantization and pruning~\citep{llm_compression_survey}, which restrict their performance potential~\citep{xu2024device}. 
Nonetheless, studies show they perform routine tasks---such as text summarization, intent recognition, and local query resolution---with adequate competence~\citep{team2025gemma}.
% Their reduced parameter count compared to cloud counterparts, combined with compression techniques such as quantization and pruning, inevitably limits their performance ceiling \cite{xu2024device}. 
% However, empirical evidence suggests that current on-device models already demonstrate sufficient competence for routine tasks including text summarization, intent recognition, and local query resolution. 
However, the lack of a comprehensive evaluation leaves their true capabilities underexplored. 
As these models evolve rapidly, establishing a comprehensive evaluation framework is crucial to ensure reliability, scalability, and alignment with user needs.
% \qy{citation}

Numerous methods and readily available benchmarks have been proposed for evaluating the practical viability of LLMs for on-device deployment. 
For instance, MLPerf Client~\citep{mlcommons_mlperf_client_v06} evaluates Time to First Token (TTFT) and Tokens Per Second (TPS) for the 4-bit quantized Llama 2 model~\citep{touvron2023llama} across tasks inspired by real-world applications, such as content generation and text summarization. 
Similarly, LocalScore~\citep{pais2025localscore} assesses pre-filling throughput, text generation throughput, and TTFT for 4-bit quantized models of 1B, 8B, and 14B parameters.
Power consumption metrics have also been explored, as demonstrated by~\cite{stevens2024electricLLM} who measure consumption in Watt-hours. 

However, a common limitation is that most existing works typically concentrate on a subset of metrics, often evaluating accuracy, efficiency, or power consumption in isolation. 
While \citet{husom2025sustainable} concurrently evaluated output accuracy, inference performance, and energy efficiency, their analysis lacks a detailed examination of how these metrics are influenced by varying workloads (e.g., model size, token length) and specific methodologies (e.g., CPU operator implementations for quantization techniques).

To address these limitations, we propose a tripartite evaluation framework for on-device LLM inference that systematically considers: 
% variations in model size and workload (defined by token length). This framework encompasses: 
1) \textit{Model capability:} assessing task-specific accuracy using standardized benchmarks; 2) \textit{Deployment efficiency:} quantifying generation throughput and latency under practical hardware constraints; 3) \textit{System resource utilization:} analyzing resource consumption and potential contention, particularly in environments with concurrent applications. 
This multidimensional framework aims to provide a holistic understanding of the scenarios where, and methodologies by which, on-device LLMs can effectively serve as alternatives to cloud-based solutions.

Our investigation employs two configurations: 1) a representative consumer-grade laptop with a single CPU and 16GB RAM, mirroring typical user hardware configurations and 2) a profiled configuration with varied computational resources, modeling the scenario with heterogeneous  resources. 
Through rigorous experimentation, we address three pivotal questions: 
1) We establish the maximum model size that maintains acceptable system responsiveness during concurrent multitasking; 
2) We dissect the trade-offs between quantization precision levels (4-bit to 8-bit) across our evaluation dimensions, revealing nonlinear relationships between bit-width reduction and performance degradation; 
3) We compare different post-training quantization methods, demonstrating how algorithmic choices affect deployment outcomes. 
These findings provide actionable insights for optimizing the balance between model efficacy and resource efficiency in edge computing environments. 
Our primary contributions are summarized as follows:

\textbf{1) Systematic Evaluation Framework:}
We propose a holistic framework that evaluates model capability, deployment efficiency, and system resource utilization. This approach comprehensively characterizes the user experience of real-world, on-device LLM inference services.

\textbf{2) Comprehensive Evaluation of On-Device LLM Inference:}
We conduct an extensive empirical study of LLM inference on consumer edge devices. Our investigation encompasses eleven LLMs across diverse parameter scales, subjected to seven distinct quantization methods. We benchmark model capability across five open-source datasets, while assessing inference efficiency and resource utilization under four varying input token lengths. This methodology ensures robust coverage across diverse model architectures, quantization levels, task types, and workloads.

\textbf{3) Analytical Observations and Insights:}
We quantify the determinants of inference performance---specifically capability, efficiency, and resource consumption---across heterogeneous model scales and quantization schemes. Our results demonstrate that: 
1) Inference performance is strongly correlated with the model parameter count; 
2) BPW is not the sole determinant of performance, as computational efficiency is specific to the quantization method and relies on hardware-aware implementation; 
3) Efficiency bottlenecks shift from computation to data transfer (communication) as model sizes increase. 
Moreover, we provide actionable guidelines for deploying LLMs in production edge environments, analyzing the trade-offs between task accuracy and efficiency to assist in optimal model selection.

% We provide recommendations for architectural optimizations pertinent to LLM inference frameworks. 

% We offer insights for accelerating LLM inference on resource-constrained edge devices by analyzing potential bottlenecks, particularly the increasing communication overheads associated with model sizes.
% \begin{enumerate}
%     \item 

%     \item 

%     \item 
% \end{enumerate}

% Research questions

% \begin{itemize}
%     \item In the cases of both parameter quantization and no parameter quantization, what scale of large language model can run locally on a device with 16GB of memory?
%     \item Among the models that can run locally, what scale of large language model can operate without affecting other tasks?
%     \item For those quantized models, what is the performance gap between them and their original counterpart?
%     \item How much impact does the local running model have on the battery life of the terminal device?
% \end{itemize}

\section{Preliminaries} \label{sec:preliminary}

\textbf{LLM Inference Frameworks}. 
The domain of edge-oriented frameworks facilitating the local deployment of LLMs has witnessed rapid advancement. 
Prominent frameworks include \q{llama.cpp}~\cite{ggmlquant2023}, TensorRT-LLM~\cite{nvidia_tensorrt_llm_2023}, SGLang~\cite{zheng2024sglang}, ExLlama~\cite{exllama2023}, and vLLM~\cite{kwon2023efficient}. 
However, frameworks such as TensorRT-LLM, SGLang, and ExLlama are primarily optimized for GPU architectures. 
Regarding vLLM, our experiments on Windows systems (via WSL) show suboptimal performance on CPU backends: we observed unacceptably low throughput (e.g., 3.07 token per second (TPS) for a 0.5B model) and encountered Out-Of-Memory errors with models exceeding 1.5B parameters. 
Furthermore, the vLLM CPU backend currently lacks support for \q{int4}/\q{int8} quantization. 
Similarly, MLC-LLM yielded unsatisfactory results, characterized by high latency and inference stalling after limited token generation. 
Given the lack of viable alternatives, we selected \q{llama.cpp} to investigate the performance limits of LLMs on consumer hardware. 
This choice is further justified by its status as the foundational backend for widely adopted tools such as Ollama~\cite{ollama2023}.

\textbf{LLM Quantization.} 
Deploying LLMs on resource-constrained devices requires model compression~\citep{llm_compression_survey}, with post-training quantization being a leading strategy due to its ability to reduce model size and computational cost with minimal performance impact.
% Post-Training Quantization (PTQ) is particularly relevant for on-device applications due to its simplicity and efficiency.
A primary framework categorizes quantization methods into \textit{symmetric} and \textit{asymmetric} variants based on their alignment with the origin. 
Following \cite{llamacppquants2024}, let $x \in \mathbb{R}$ be an original (pre-quantization) value and $q$ its quantized counterpart, using $n \in \mathbb{N}^+$ bits of precision, symmetric quantization employs $n$-bit signed integers, where the discrete quantized value $q$ lies in the range $\{-2^{n-1}, \dots, 2^{n-1}-1\}$. 
A scaling factor $s \in \mathbb{R}$ is computed as $s = \frac{|x|_{\max}}{2^{n-1}}$, where $|x|_{\max}$ is the maximum absolute value of $x$.
Denoting rounding by $\text{Round}(\cdot)$, 
% $\nint{\cdot}$, 
$q$ is computed
% obtained by scaling, rounding, and clamping 
as $q = \max\left(-2^{n-1}, \min\left(\text{Round}\left( \tfrac{x}{s} \right), 2^{n-1}-1\right)\right).$
% Consequently, the original and quantized values share a common origin.
Asymmetric quantization employs $n$-bit unsigned integers, resulting in $q \in \{0, \dots, 2^n-1\}$.
Let $m = x_{\min}$ and $x_{\max}$ be the minimum and maximum values of the input data $x$, respectively.
The scaling factor $s$, which maps the input range $[m, x_{\max}]$ to approximately $[0, 2^n-1]$, is computed as $s = \frac{x_{\max} - m}{2^n - 1}$. 
The input $x$ is then quantized as $q = \max\left(0, \min\left(\text{Round}(\frac{x-m}{s}), 2^n-1\right)\right)$.
% to $q$ by applying a similar offset, scaling, rounding, and clamping as $q = \max\left(0, \min\left(\nint{\frac{x-m}{s}}, 2^n-1\right)\right)$.

\textbf{Quantization Methods in \q{llama.cpp}.}
\q{llama.cpp}~\citep{llamacppquants2024} stands out as a crucial tool in the landscape of on-device LLM research and development due to its open-source and cross-platform nature.
The quantization schemes \q{n\_k} implemented in \q{Llama.cpp} is adopted in our study.
% The \texttt{n\_0} method employs symmetric quantization, 
% % with a fixed block size of 32 (\textit{$d_1$} = 32), 
% where weights are scaled uniformly using a zero-centered range, eliminating the need for zero-point offsets.
% In contrast, 
The \q{n\_K} framework uses asymmetric strategy and further extends through five synergistic enhancements: 
1) Hierarchical parameter grouping, which recursively quantizes scale $s$ and offset $m$ metadata to reduce overhead; 
2) Activation-guided importance matrices that prioritize high-impact dimensions during quantization, mitigating accuracy loss from skewed weight distributions; 
3) Post-quantization convex optimization to refine scales and zero-points by minimizing layer-wise reconstruction error; 
4) Perturbative search algorithms that iteratively adjust quantized values to escape local minima, improving parameter recovery; 
5) Heterogeneous bit allocation, assigning different precision to different weight matrices.

% This simplicity ensures computational efficiency, particularly on mobile CPUs with limited parallelization capabilities \cite{ggmlquant2023}. 
\section{Evaluation Framework}

This section presents our proposed framework to systematically evaluate on-device LLMs. We first describe our selection criteria for quantization methods and models. Then, we present the evaluation framework, evaluating 1) model capability, 2) deployment efficiency, and 3) system resource utilization, respectively.

% \qy{NOTE: move configurations to experiments}

\subsection{Model and Quantization Selection} \label{sec:models}
% \begin{itemize}
%     \item Llama 3 includes 3.1-8B, 3.1-405B, 3.2-1B, 3.2-3B, 3.3-70B
%     \item Qwen2.5 includes 0.5B, 1.5B, 3B, 7B, 14B, 32B, and 72B;
%     \item Deepseek includes V3 and R1. Both have 671B parameters
%     % \item Mistral () includes 7B-Instruct-v0.2, 8$\times$7B-Instruct-v0.1, Small-Instruct-2409 (todo)
% \end{itemize}
% \textcolor{RedOrange}{(qingyu)}
% (introduce models used in each research question and how they are quantized (if needed))
We consider the Qwen 2.5~\citep{qwen2024qwen2.5}, Qwen 3~\citep{qwen2025qwen3}, and Llama 3~\citep{llama2024llama3} series of LLMs due to their widespread adoption and popularity within the research community and industry. 
Specifically, Qwen 2.5 models with 0.5B, 1.5B, 3B, 7B, 14B parameters, Qwen 3 models with 0.6B, 1.7B, 4B, 8B parameters, Llama 3 models with 1.5B, 3B, 8B parameters are selected for evaluation. 
The upper limit of 14B parameters was chosen as it represents the largest model size that can be reliably deployed on laptop with 16GB RAM after quantization.
% The 14B upper bound is determined by post-quantization memory viability: a 14B model quantized to 4-bit precision with auxiliary metadata (scales, offsets) requires 10.2GB RAM \cite{dettmers2023qlora}, aligning with the 16GB threshold of mainstream laptops.

For effective on-device LLM deployment, both the quantization method and the resulting data format are critical considerations. 
The choice of quantization method, such as symmetric or asymmetric quantization, block-wise or per-tensor quantization, directly influences the trade-off between model size reduction and potential accuracy loss. 
The data format resulting from quantization, which dictates the bit-width and representation of the quantized weights, significantly impacts memory footprint and computational efficiency. Thus, considering the availability in \q{llama.cpp}, five quantization methods i.e., \q{q8\_0}, \q{q5\_k}, \q{q4\_k}, \q{q3\_k}, and \q{q2\_k}, are adopted in this study. The effective bit-widths are about 8 bits, 5 bits, 4 bits, 3 bits, 2 bits, respectively.

\subsection{Model Capability Evaluation}
\label{subsec:task-performace}

\textbf{Datasets.}
We have selected five diverse benchmarks that encompass a broad spectrum of competencies. GSM8K~\citep{cobbe2021training} emphasizes multi-step reasoning by presenting grade-school mathematics problems. HellaSwag~\citep{zellers2019hellaswag} evaluates commonsense reasoning, challenging models to infer plausible continuations in everyday scenarios. Similarly, MMLU~\citep{hendrycks2021mmlu} spans a wide array of disciplines, providing an extensive evaluation of general knowledge. Together, these benchmarks facilitate a comprehensive assessment of a model's performance, resilience, and adaptability under resource constraints. HumanEval~\citep{chen2021evaluating} assesses the model's capabilities in generating precise and efficient code, reflecting practical software development tasks. Lastly, TruthfulQA~\citep{lin2022truthfulqa} examines the reliability of factual outputs and mitigates the risks related to hallucinated information.

\textbf{Evaluation Metrics.}
Evaluation metrics for the aforementioned tasks must be tailored to their unique characteristics to ensure accurate and meaningful assessment of model performance. 
Each task presents distinct challenges and objectives, necessitating the use of specialized metrics that align with its specific requirements. 
We adopt \q{lm-evaluation-harness}~\citep{gao2024evalharness} (version \q{0.4.8}) as the evaluation tool to align with Open-LLM-Leaderboard so that all our evaluations and comparisons are fair and clear.
Given the incomplete native support of \q{lm-evaluation-harness} for \q{llama.cpp}, we develop a customized evaluation mechanism, which extends \q{lm-evaluation-harness} to ensure comprehensive functionality.
We keep the default settings on most hyper-parameters such as temperature and top-$p$ threshold. We set the same number of few shots according to the Qwen 2.5 technical report~\citep{qwen2024qwen2.5}: 4-shots for GSM8K, 10-shots for HellaSwag, 5-shots for MMLU, and 0-shot for TruthfulQA and HumanEval.

\subsection{Deployment Efficiency Evaluation}
\label{sec:deployment_efficiency}
\textbf{Datasets.} 
To evaluate efficiency and sustained operational performance, we conduct synthetic dataset with fixed input and output lengths of generated tokens. The configurations for input and output token-length are \{64, 128, 256, 512\} and \{1024\}, respectively. 
To approximate real-world application scenarios, initial inputs consist of text segments with lengths of 54, 118, 246, or 502 tokens. 
Each segment is prefixed with a constant 10-token (counted by \q{Llama.cpp}) instructional prompt (\textit{Write a 50000-word article based on this text:}) to elicit 50000 tokens generation. Then, we uniformly terminate the generation process after 1024 tokens. Moreover, the synthetic input tokens are preliminarily generated by a LLM.

\textbf{Evaluation Metrics.}
The primary metric for LLM inference efficiency is throughput, measured in TPS. We evaluate throughput independently for the prefill (input processing) and decode (token generation) stages, defining it as the number of tokens processed divided by the execution time. For our experiments, all performance metrics are obtained directly from the runtime logs provided by the \q{Llama.cpp} framework. We have verified the accuracy of these logs; our manual measurements of latency and throughput align with the logged values at three decimal places.

To ensure reliable measurements, we first pre-warm the \q{Llama.cpp} with three preliminary runs to mitigate cold-start latency. And we fix \q{temperature} and \q{seed} parameters to ensure reproducibility. In the event that a run fails to produce the target 1024 tokens, these parameters are adjusted, and the trial is repeated. We calculate mean value of three independent runs, with the KV-cache cleared before each trial to ensure measurement independence.

We establish two experimental configurations to comprehensively evaluate throughput:

1) \textbf{Real-World Consumer Devices:} A laptop-based setup with fixed computational resources and memory specifications.

2) \textbf{Profiled Testbed:} A profiled environment utilizing heterogeneous resource allocations (varying CPU cores and execution allowances). It emulates job preemption and resource contention typical of real-world deployments. Specifically, we configure the CPU core allocation as $\{4, 8, 12\}$ and the CPU execution allowance as $\{50\%, 75\%, 100\%\}$ of the maximum capacity.

% \paragraph{Evaluation Method.}
% We calculate the mean value of throughput results over three independent runs to ensure reliability and clear the KV-cache after each run.
% We utilize \q{llama-cpp-python}~\cite{abetlen2025llamacpppython}, the open-source Python binding which operates as a Python interface for the underlying \texttt{llama.cpp} framework.
% All reported experimental results are derived from the invocation of the \q{\_\_call\_\_} function. To mitigate performance fluctuations caused by varying token-generation lengths, we standardize the number of generated tokens across all models using a fixed instructional prompt and the parameter \q{max\_token=1024}. 
% We calculate the mean value of throughput results over three independent runs to ensure reliability. After each execution, the KV-cache is explicitly cleared by invoking the \q{reset} function provided by \q{llama-cpp-python}.
% We calculate the mean value of throughput results over three independent runs to ensure reliability. After each execution, the KV-cache is explicitly cleared by invoking the \q{reset} function provided by \q{llama-cpp-python}.

\subsection{System Resource Utilization Evaluation} 
\label{sec:system_resource_utilization}
\textbf{Datasets.}
Following the setup in \Cref{sec:deployment_efficiency}, we evaluate the performance of token generation with input lengths of 64, 128, 256, and 512 tokens.

\textbf{Evaluation Metrics.} 
We evaluate the resource utilization for Windows system laptop and design three primary metrics: CPU utilization, memory occupancy, and power consumption. 

First, we utilize Windows Performance Recorder (WPR) and  to systematically quantify CPU power throughout ten executions for every model with each quantization method. WPR excludes power cost for data communication with RAM~\cite{Microsoft_WPR}. We analyze the results with Windows Performance Analyzer (WPA)~\cite{Microsoft_WPA}. 

Then, the \q{psutil} Python library~\cite{psutil} is employed to quantify memory consumption. This library facilitates the retrieval of system and process information, including detailed memory usage statistics. 
Our analysis focuses on the Resident Set Size (RSS), representing the non-swapped physical memory utilized by the LLM inference process~\cite{psutil_doc}. 
The key metric adopted is Peak RSS, defined as the maximum physical RAM consumed by the process at any point during its execution, thereby capturing the worst-case memory footprint. 
We select the ``wset'' metric by \q{psutil}, aligning with the ``Memory (Working Set)'' in Windows task manager, consistent with system-level reporting.

Moreover, we utilize Windows Energy Estimation Engine (E3) to collect energy consumption. E3 is an integrated real-time power modeling system, which translates hardware activity metrics into energy estimates for various system components~\cite{e3_meansure2020}. 
We implement a customized WPR profile through a structured XML configuration to optimize data collection and minimize extraneous recording, thereby logging target energy-relevant counters.

\section{Experiments} \label{sec:experiments}

Our experiments utilize \q{llama-cpp-python}~\citep{abetlen2025llamacpppython}, an open-source Python binding for \q{llama.cpp}. The evaluation platform is a consumer-grade laptop with a 12-core, 2.20 GHz Core i7-1360P CPU, 16 GB of 5600 MT/s DDR5 RAM, and a NVMe SSD, selected to represent a typical user configuration. And a Linux server with Ubuntu 22.04 on a 16-core 2.40 GHz EPYC 9654 CPU and 64 GB of 4800 MT/s DDR5 RAM to profile a testbed with heterogeneous computational resources. Both CPUs supports Advanced Vector Extensions (AVX2) instruction extension~\cite{intel_sdm}.

We compile \q{llama.cpp} on both platforms and implement the testbed with Docker container (version 29.1.3). We use \q{--cpus} option to configure CPU cores and use \q{--cpuset-cpus} to set CPU quotas~\cite{docker_resource_constraints_2026}. 
Moreover, since the model capability performance is platform-independent~\citep{schlogl2023causes}, we conducted all performance evaluations on a GPU to achieve acceleration. These evaluations were performed on a Linux workstation equipped with two 56-core server-class CPUs and a data-center GPU with 80 GB of VRAM.

%  (v\q{0.3.7})

\subsection{Model Capability Results}
\label{sec:results_model_capability}

Following the methodology outlined in \Cref{subsec:task-performace}, we assess the capabilities of representative LLMs, i.e., Qwen 2.5 and Llama 3, across five standard open-source benchmarks: GSM8K, HellaSwag, MMLU, HumanEval, and TruthfulQA. The results are shown in \Cref{tab:task-qwen2.5-performace,tab:task-llama3-performace}. Due to the page limit, the detailed analysis for Llama results can be found in \Cref{sec:additional_results_model_capability}.
Based on the empirical results, we derive two primary observations regarding model capability:

1) Model capability generally deteriorates as model sizes and BPW decreases. Notably, extreme compression (e.g., 2-bit quantization) results in critical performance loss.

2) Large models demonstrate superior resilience to the performance degradation induced by aggressive low-precision quantization compared to smaller models.

First, we examine the impact of parameter scaling and quantization.
For a fixed quantization method, models with larger parameter counts consistently demonstrate superior accuracy across all benchmarks, indicating enhanced capacity for reasoning, knowledge retention, and generalization. 
Specifically, the 14B models outperform all smaller variants, securing top scores on GSM8K (\q{q4\_0}), HellaSwag (\q{q5\_k}), MMLU (\q{q5\_0}), and TruthfulQA (\q{q5\_0}). 
Conversely, smaller architectures (e.g., the 3B and 1.5B variants) exhibit significant accuracy degradation on reasoning-intensive benchmarks such as GSM8K and MMLU. 
Notably, the 7B models---particularly the \q{q8\_0} variant---achieve performance on HumanEval comparable to, or marginally exceeding, that of the 14B models. 
This phenomenon is consistent with findings reported in~\cite{qwen2024qwen2.5}.

For a fixed model size, the results indicate clear performance degradation attributable to reduced precision. 
As detailed in \Cref{tab:task-qwen2.5-performace}, the \q{fp16} configuration generally yields optimal accuracy. 
For instance, the 1.5B model achieves 60.80\% accuracy on GSM8K with \q{fp16}, whereas performance declines significantly to 52.77\% when using \q{q4\_k}. Extreme quantization (\q{q2\_k}) degrades performance severely across all sizes, compromising both accuracy and fidelity.
\begin{table}[tbp]
  \centering
  \caption{Capability Benchmarks of Selected Qwen 2.5 Models}
  \vspace{-4mm}
  \label{tab:task-qwen2.5-performace}
  \renewcommand{\arraystretch}{0.8}
  \resizebox{\linewidth}{!}{%
  \begin{tabular}{c|l|*{9}{c}}
    \toprule
    \multicolumn{2}{c|}{\multirow{2}{*}{\textbf{Models \& Tasks}}} &\multicolumn{8}{c}{\textbf{Quantization Methods (High BPW $\to$ Low BPW)}} \\
    \multicolumn{2}{c|}{} &\q{fp16} &\q{q8\_0} &\q{q5\_k} &\q{q5\_0}  &\q{q4\_k} &\q{q4\_0} &\q{q3\_k} &\q{q2\_k} \\
    \midrule
    \multirow{5}{*}{14B}
    &GSM8K~\citep{cobbe2021training}        
      &/ &/ &89.08 &89.76 &88.86 &\textbf{90.45} &89.01 &84.46 \\
    &HellaSwag~\citep{zellers2019hellaswag} 
      &/ &/ &\textbf{85.09} &84.78 &84.72 &84.24 &83.93 &81.76 \\
    &MMLU~\citep{hendrycks2021mmlu}         
      &/ &/ &79.74 &\textbf{79.75} &79.55 &79.43 &78.86 &75.74 \\
    &HumanEval~\citep{chen2021evaluating}   
      &/ &/ &\textbf{69.51} &\textbf{69.51} &68.90 &68.90 &\textbf{69.51} &62.20 \\
    &TruthfulQA~\citep{lin2022truthfulqa}   
      &/ &/ &68.64 &\textbf{69.63} &68.93 &67.50 &66.92 &65.89 \\
    \midrule
    \multirow{5}{*}{7B}
    &GSM8K~\citep{cobbe2021training}        
      &/ &86.73 &86.05 &\textbf{87.11} &85.52 &86.20 &84.46 &76.19 \\
    &HellaSwag~\citep{zellers2019hellaswag} 
      &/ &\textbf{81.32} &81.21 &81.19 &80.94 &80.59 &79.79 &77.48 \\
    &MMLU~\citep{hendrycks2021mmlu}         
      &/ &74.24 &\textbf{74.27} &74.15 &74.26 &74.04 &73.23 &68.58 \\
    &HumanEval~\citep{chen2021evaluating}   
      &/ &\textbf{70.73} &67.68 &68.90 &64.63 &58.54 &63.41 &53.66 \\
    &TruthfulQA~\citep{lin2022truthfulqa}   
      &/ &\textbf{64.74} &64.25 &64.45 &63.80 &62.22 &64.44 &62.25 \\
    \midrule
    \multirow{5}{*}{3B}
    &GSM8K~\citep{cobbe2021training}        
      &\textbf{80.89} &80.29 &78.77 &80.82 &76.80 &75.36 &62.62 &48.52 \\
    &HellaSwag~\citep{zellers2019hellaswag} 
      &\textbf{75.29} &75.07 &74.90 &75.10 &74.86 &73.69 &70.85 &68.56 \\
    &MMLU~\citep{hendrycks2021mmlu}         
      &66.38 &\textbf{66.52} &66.06 &66.34 &65.68 &65.08 &60.15 &58.80 \\
    &HumanEval~\citep{chen2021evaluating}   
      &\textbf{54.88} &\textbf{54.88} &49.39 &47.56 &48.78 &45.12 &46.34 &31.10 \\
    &TruthfulQA~\citep{lin2022truthfulqa}   
      &58.67 &58.59 &58.38 &\textbf{58.94} &58.38 &57.00 &56.38 &50.16 \\
    \midrule
    \multirow{5}{*}{1.5B}
    &GSM8K~\citep{cobbe2021training}        
      &\textbf{60.80} &59.82 &59.14 &57.85 &52.77 &53.30 &46.47 &21.46 \\
    &HellaSwag~\citep{zellers2019hellaswag} 
      &67.91 &\textbf{67.92} &67.70 &67.66 &66.81 &66.89 &65.06 &59.92 \\
    &MMLU~\citep{hendrycks2021mmlu}         
      &60.28 &\textbf{60.36} &60.32 &59.77 &59.76 &59.14 &57.33 &51.30 \\
    &HumanEval~\citep{chen2021evaluating}   
      &37.20 &37.80 &34.76 &\textbf{39.02} &37.50 &35.37 &24.39 &20.73 \\
    &TruthfulQA~\citep{lin2022truthfulqa}   
      &46.70 &46.70 &45.78 &46.41 &45.41 &\textbf{47.71} &45.21 &46.51 \\
    \midrule
    \multirow{5}{*}{0.5B}
    &GSM8K~\citep{cobbe2021training}        
      &31.48 &\textbf{33.28} &31.77 &30.86 &30.33 &21.00 &25.85 &21.99 \\
    &HellaSwag~\citep{zellers2019hellaswag} 
      &\textbf{50.50} &50.37 &49.88 &50.12 &50.20 &48.52 &49.65 &49.04 \\
    &MMLU~\citep{hendrycks2021mmlu}         
      &46.69 &\textbf{46.81} &46.09 &46.08 &46.12 &44.36 &45.53 &44.83 \\
    &HumanEval~\citep{chen2021evaluating}   
      &29.88 &\textbf{31.71} &26.83 &29.27 &28.66 &22.56 &28.05 &25.00 \\
    &TruthfulQA~\citep{lin2022truthfulqa}   
      &42.50 &42.50 &42.15 &\textbf{42.61} &42.51 &40.40 &41.89 &40.90 \\
    \bottomrule
  \end{tabular}
  }%
\end{table}

Second, larger models display superior stability across quantization levels relative to smaller models. 
On the GSM8K benchmark, the 14B model maintains performance, shifting only from 89.08\% (\q{q5\_k}) to 89.01\% (\q{q3\_k}). 
Conversely, the 1.5B model shows significant sensitivity, with accuracy falling from 59.14\% (\q{q5\_k}) to 46.47\% (\q{q3\_k}). 
This disparity suggests that while large models tolerate aggressive quantization, smaller models require careful precision management to remain viable for on-device applications.

Additionally, task-specific evaluations reveal critical dependencies on model capacity. 
Tasks involving code generation, such as HumanEval, prove particularly sensitive to compression. 
Notably, the 1.5B \q{q5\_0} model achieves 39.02\% on HumanEval, significantly lagging behind the 14B model's 69.51\%. 
A parallel trend is observed in TruthfulQA, where factual reliability declines alongside model size. 
These results highlight the substantial challenge of maintaining high-fidelity performance in smaller, quantized architectures.

\subsection{Deployment Efficiency Results}
\label{sec:results_deployment_efficiency}

Following the design in \Cref{sec:deployment_efficiency}, we evaluate the prefilling and decoding efficiencies of representative LLMs across diverse quantization methods. Our analysis highlights the impact of model size, input token length, computational resource allocation, and quantization schemes on throughput. We derive three key observations:

1) Throughput declines monotonically as model scale increases.

2) The primary performance bottleneck shifts with model size: large models are predominantly communication-constrained, whereas small models are compute-constrained.

3) Despite higher computational complexity, \q{QN\_k} methods can surpass \q{QN\_0} with hardware-specific enhancements.

\begin{figure}[t]
  \centering

  \begin{subfigure}{\columnwidth}
    \centering
    \includegraphics[width=\linewidth]{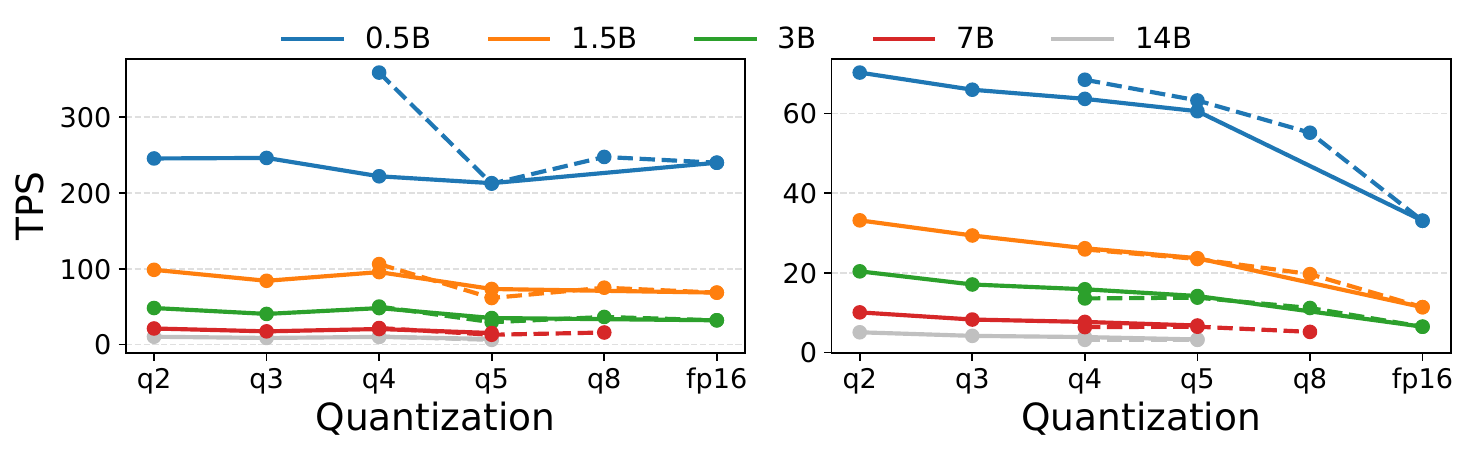}
    \vspace{-5mm}
    \caption{Qwen: Prefilling (Left), Decoding (Right)}
  \end{subfigure}
  
  \begin{subfigure}{\columnwidth}
    \centering
    \includegraphics[width=\linewidth]{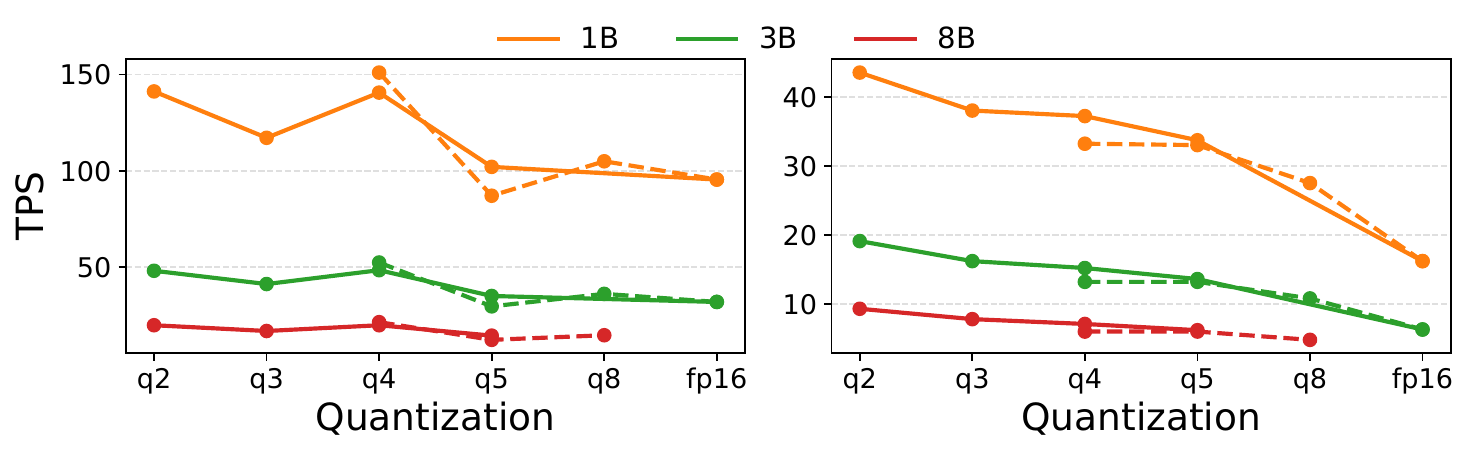}
    \vspace{-5mm}
    \caption{Llama: Prefilling (Left), Decoding (Right)}
  \end{subfigure}
  \vspace{-7mm}
  \caption{Throughput across Quantization Methods (Solid: \q{QN\_k}, Dahsed: \q{QN\_0})}
  \label{fig:throughput_128token}
\end{figure}

\begin{figure}[t]
  \centering
  \begin{subfigure}{\columnwidth}
    \centering
    \includegraphics[width=\linewidth]{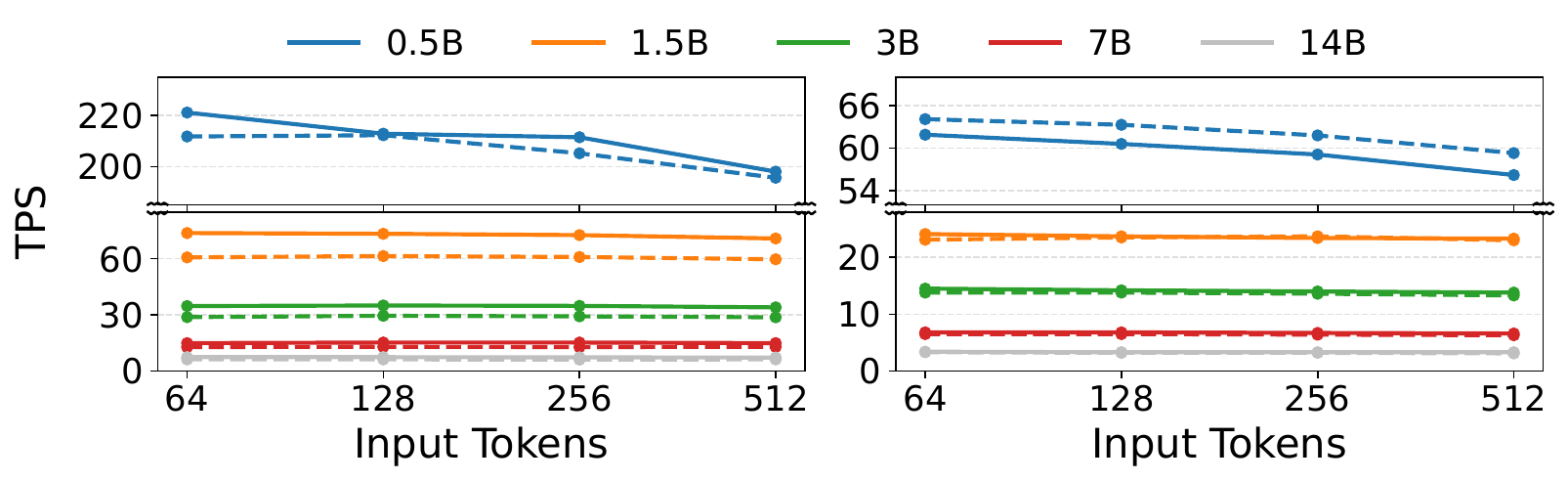}
    \vspace{-5mm}
    \caption{Qwen: Prefilling (Left), Decoding (Right)}
  \end{subfigure}
  
  \begin{subfigure}{\columnwidth}
    \centering
    \includegraphics[width=\linewidth]{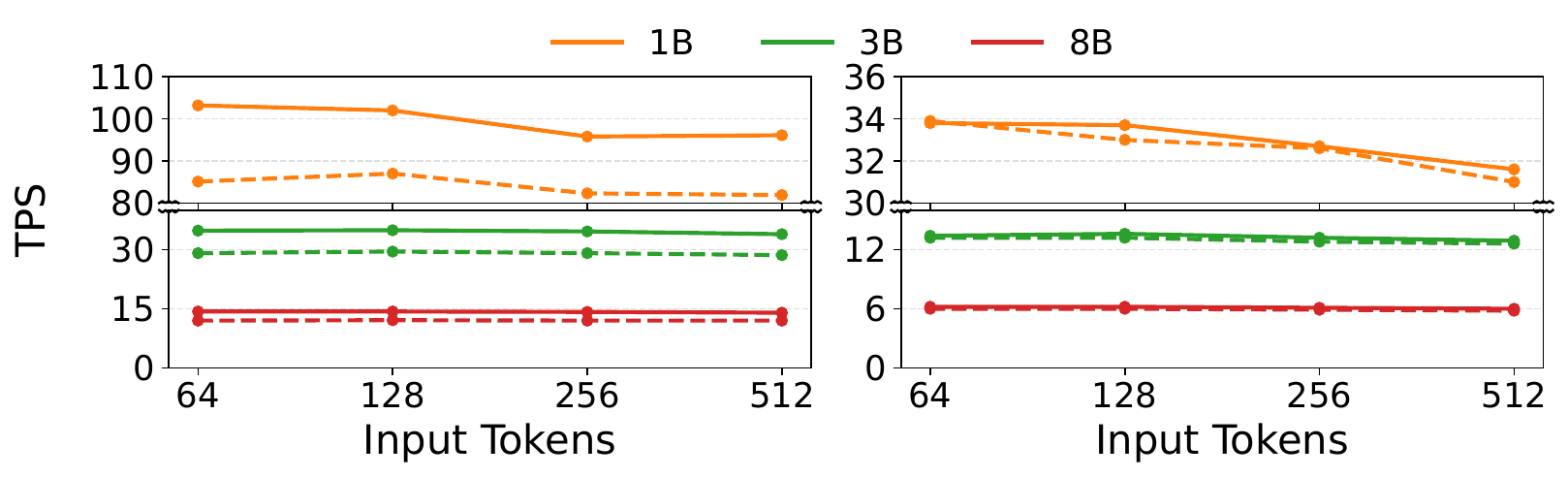}
    \vspace{-5mm}
    \caption{Llama: Prefilling (Left), Decoding (Right)}
  \end{subfigure}
  
  \vspace{-3mm}
  \caption{Throughput across Token Lengths (Solid: \q{Q5\_k}, Dahsed: \q{Q5\_0})}
  \label{fig:token-throughput-q5}
\end{figure}

With a fixed output length of 1024 tokens, we first evaluate throughput across varying quantization methods using a fixed input length of 128 tokens. Subsequently, we assess the throughput of the \q{Q5} methods (\q{Q5\_k} and \q{Q5\_0}) across diverse input lengths. The results are respectively illustrated in \Cref{fig:throughput_128token,fig:token-throughput-q5}, where solid bold curves denote the \q{QN\_k} quantization family, while dashed curves represent the \q{QN\_0} family. Individual model architectures, as detailed in \Cref{sec:models}, are distinguished by color. In \Cref{fig:token-throughput-q5}. The results show that throughput declines monotonically with increasing BPW in both the prefilling and decoding phases, a result attributed to the elevated computational and communication overheads.

\textbf{Throughput Bottleneck Identification}. Further, we characterize the system's performance limitations by analyzing throughput degradation ratios attributable to computation and communication. Given that the prefilling phase is inherently compute-bound while decoding is communication-bound.

We define the operational intensity as FLOPs/Bytes, which characterizes the ratio of computation to communication. To isolate these factors, we design two experimental scenarios:
\textit{Case 1 (Communication Impact):} We evaluate throughput degradation caused by communication overhead by comparing \q{Q2\_k} and \q{FP16} performance. We calculate the degradation ratio as $(T_{\text{\q{Q2\_k}}} - T_{\text{\q{FP16}}}) / T_{\text{\q{Q2\_k}}}$. Transitioning from 2-bit to 16-bit precision results in an approximately $8\times$ increase in communication volume (decoding) while reducing the operational intensity to roughly $1/8$ (prefilling).
\textit{Case 2 (Computation Impact):} We evaluate throughput degradation driven by computation overhead by varying the input length from 64 to 512 tokens. The degradation ratio is calculated as $(T_{64} - T_{512}) / T_{64}$. This scenario imposes an $8\times$ increase in computational load during prefilling. Conversely, the communication overhead during decoding remains largely unchanged, as the incremental communication cost of the KV-cache is negligible relative to that of the model weights.
The results are presented in \Cref{tb:degrade_ratios}.
\begin{table}[t]
\centering
\caption{Throughput Degradation Ratios across Varying BPW and Input Sequence Lengths.}
\vspace{-4mm}
\label{tb:degrade_ratios}
\resizebox{\linewidth}{!}{
\begin{tabular}{@{}l cc cc@{}}
\toprule
\multirow{2}{*}{\makecell[c]{Model}}
 & \multicolumn{2}{c}{(\q{Q2\_k} - \q{FP16})/\q{Q2\_k}}
 & \multicolumn{2}{c}{(64 tokens - 512 tokens)/64 tokens} \\
\cline{2-5}
 & Prefilling (\%) & Decoding (\%)
 & Prefilling (\%) & Decoding (\%) \\
\hline
Qwen2.5-0.5B & 2.28\%  & 52.92\% & 10.49\% & 9.21\% \\
Qwen2.5-1.5B & 30.53\% & 65.66\% & 3.93\%  & 3.32\% \\
Qwen2.5-3B   & 33.61\% & 68.14\% & 2.02\%  & 4.83\% \\
Qwen2.5-7B   & /       & /       & 0.00\%  & 2.94\% \\
Qwen2.5-14B  & /       & /       & 2.70\%  & 2.94\% \\
Llama3.2-1B  & 32.44\% & 62.76\% & 6.88\%  & 6.51\% \\
Llama3.2-3B  & 33.75\% & 67.02\% & 2.59\%  & 3.73\% \\
Llama3.1-8B  & /       & /       & 2.10\%  & 3.23\% \\
\bottomrule
\end{tabular}
}
\end{table}

First, for large models (defined here as $>0.5$B parameters), Case 1 leads to significant throughput deterioration, exceeding 30\% in both the prefilling and decoding phases. In contrast, Case 2 results in only marginal degradation, remaining below 7\% for both phases. These results indicate that communication overhead is the primary bottleneck for large models. 
Conversely, for tiny models ($0.5$B parameters), the deterioration caused by Case 1 is alleviated, dropping to 5.29\% in prefilling and 2\% in decoding. However, the degradation attributed to Case 2 rises to 10.49\% in prefilling and 9.21\% in decoding. Thus, computation overhead exerts a more pronounced impact on the throughput of small models. 
These findings suggest that prioritizing computational resources for small models and communication bandwidth for large models is the most economical resource allocation strategy.

\begin{figure}[t]
  \centering
  \setlength{\abovecaptionskip}{2pt}
  \setlength{\belowcaptionskip}{0pt}

  \begin{subfigure}{\columnwidth}
    \centering
    \includegraphics[width=\linewidth]{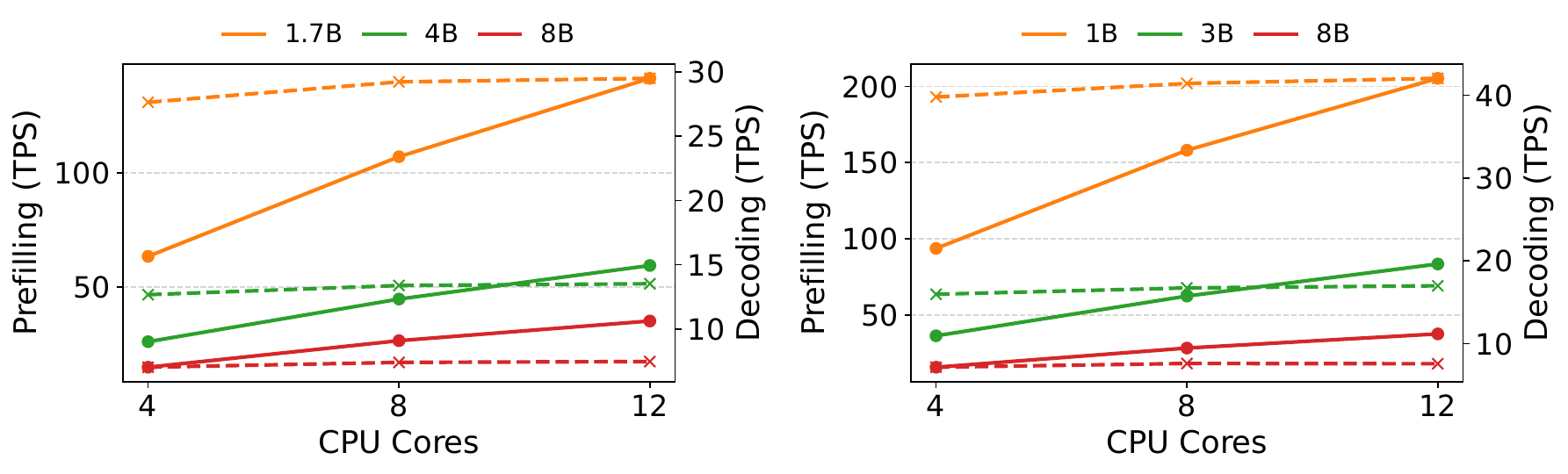}
    \vspace{-5mm}
    \caption{Throughput across CPU cores (Qwen: Left, Llama: Right)}
    \label{fig:throughput_profiled_cpu_core}
  \end{subfigure}

  \begin{subfigure}{\columnwidth}
    \centering
    \includegraphics[width=\linewidth]{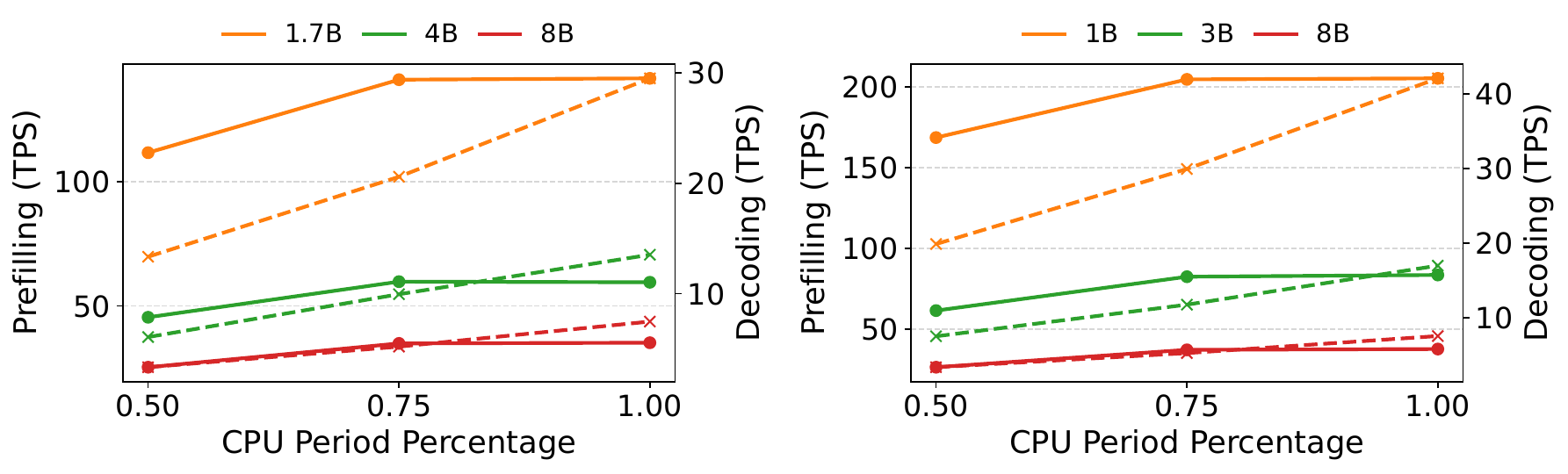}
    \vspace{-5mm}
    \caption{Throughput across CPU Runtime (Qwen: Left, Llama: Right)}
    \label{fig:throughput_profiled_cpu_runtime}
  \end{subfigure}
  
  \caption{Throughput across CPU Resources (Solid: Prefilling, Dahsed: Decoding)}
  \label{fig:throughput_profiled_cpu}
\end{figure}

\textbf{Results with Different Computational Resources}.
Moreover, we evaluate throughput scalability across varying computational resource configurations. In this experiment, we utilize the Qwen 3 and Llama 3 model families, fixing the quantization method at \q{Q4\_K} and the input length at 128 tokens to isolate the impact of resource variability. We measure throughput fluctuations across different CPU core counts with fixed 100\% CPU runtime and CPU execution quotas with 12 CPU cores.
The results are presented in \Cref{fig:throughput_profiled_cpu}, where solid and dashed lines denote prefilling and decoding throughput, respectively. Note that prefilling and decoding are plotted against separate vertical axes (left and right).
First, as illustrated in \Cref{fig:throughput_profiled_cpu_core}, increasing the number of CPU cores results in a linear increase in prefilling throughput, whereas decoding throughput exhibits only marginal gains. This observation aligns with the identification of prefilling as compute-bound and decoding as communication-bound.

Second, \Cref{fig:throughput_profiled_cpu_runtime} shows that increasing the CPU quota yields distinct throughput patterns. In the setup, CPU execution time is governed by Docker's CPU quota mechanism~\cite{docker_resource_constraints_2026}, which allocates a percentage of the total runtime within a predefined period via the Linux Completely Fair Scheduler~\cite{cfs_cpu_quota}. The resulting process throttling enforced by the Linux kernel~\cite{cpu_throttling_2022} induces frequent context switches and cache misses, significantly degrading data locality. Thus, the communication-bound decoding phase is highly sensitive to CPU quota restrictions due to these efficiency penalties. 

Moreover, as illustrated in \Cref{fig:throughput_profiled_cpu_core}, the rate of increase in prefilling throughput diminishes as model size increases. This trend indicates that for larger models, raw computational power becomes a less dominant factor in driving performance gains. Instead, it suggests that the communication bottleneck increasingly constrains the system, identifying it as the primary cause of throughput saturation in the large-model regime.

% Similarly, the reduced incline rates in decoding throughput increase in \Cref{fig:throughput_profiled_cpu_runtime} also implies the computation is the critical bottleneck for small models

\textbf{Results on Different Hardware Supports}.
Observing the distinct throughput behaviors of the \q{Q4} and \q{Q5} families (\q{Q4\_0} vs. \q{Q4\_k} and \q{Q5\_0} vs. \q{Q5\_k}) in \Cref{fig:throughput_128token}, we proceed to analyze their performance under heterogeneous hardware acceleration.

In \q{llama.cpp}, matrix operations comprise two distinct phases: unpacking and computation~\cite{llama_cpp_2024_quantc}. The unpacking phase upscales low-precision quantized weights into higher-precision formats, loading them into CPU vector registers. The target precision within these registers is dictated by the specific instruction set architecture available on the hardware. For instance, if a CPU supports Vector Neural Network Instructions (VNNI), it natively enables efficient 8-bit integer matrix multiplication~\cite{intel_avx512_dl_boost}. Notably, within the \q{llama.cpp} framework, both the \q{QN\_k} and \q{QN\_0} quantization families leverage AVX2 acceleration to optimize these operations~\cite{llama_cpp_2024_quantc}.

% However, \q{QN\_k} methods suffers lower degree of parallelism than \q{QN\_0} series due to the two-level scaling parameter \cite{llama_cpp_2024_q40_q80, llama_cpp_2024_q50_q80} \qy{todo: q4k q5k, two versions}, the VNNI implementations are not adopted in the official implementations \qy{todo: reference}.

\begin{figure}[t]
  \centering
  \setlength{\abovecaptionskip}{2pt}
  \setlength{\belowcaptionskip}{0pt}

  \begin{subfigure}{\columnwidth}
    \centering
    \includegraphics[width=\linewidth]{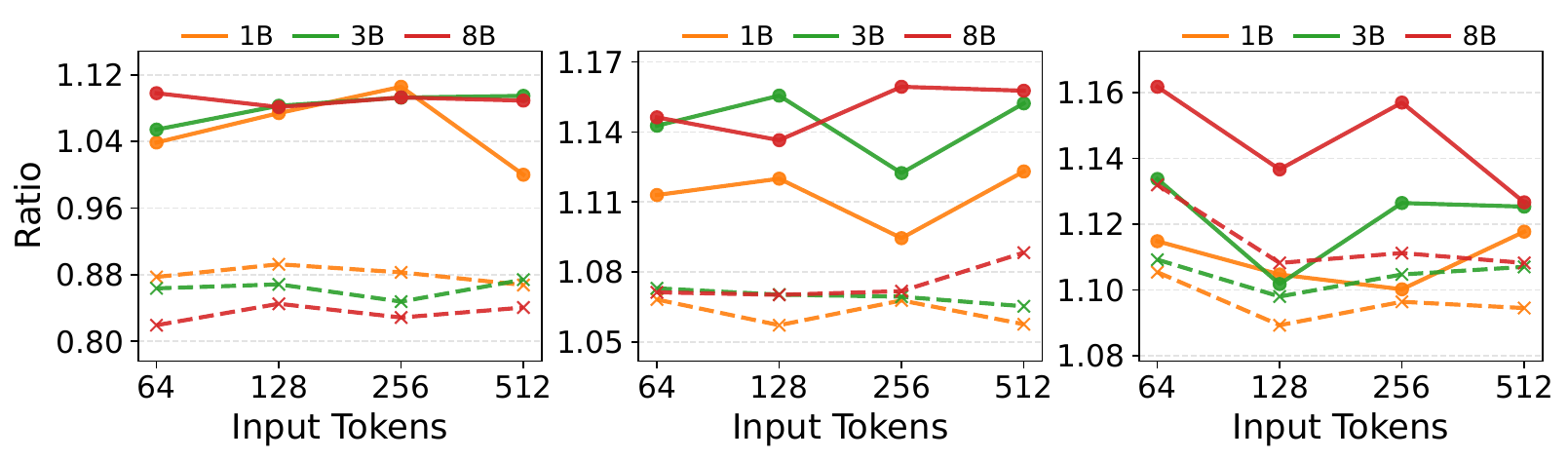}
    \vspace{-5mm}
    \caption{\q{Q4\_0}/\q{Q4\_K} Ratio}
  \end{subfigure}

  \begin{subfigure}{\columnwidth}
    \centering
    \includegraphics[width=\linewidth]{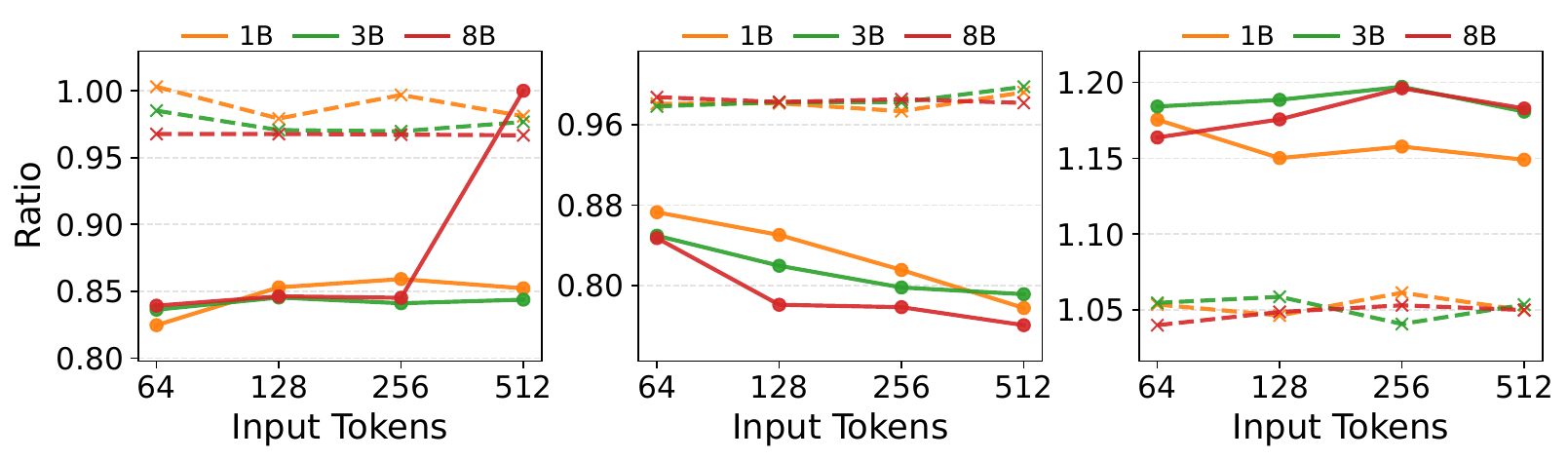}
    \vspace{-5mm}
    \caption{\q{Q5\_0}/\q{Q5\_K} Ratio}
  \end{subfigure}
  \vspace{-3mm}
  \caption{\q{QN\_0}/\q{QN\_K} Throughput Ratio across Token Lengths (Solid: Prefilling, Dahsed: Decoding), Laptop with VNNI (Left), Testbed with VNNI (Middle), Testbed w/o VNNI (Right).}
  \label{fig:ratio_q4_q5}
\end{figure}

We experimentally evaluate Llama 3 models by analyzing the throughput ratios of \q{Q4\_0} to \q{Q4\_k} and \q{Q5\_0} to \q{Q5\_k} during prefilling and decoding. As illustrated in \Cref{fig:ratio_q4_q5}, solid lines represent prefilling, while dashed lines represent decoding. The figure is segmented by hardware configuration: the left and center panels display results with AVX2-VNNI acceleration enabled (on the laptop and testbed, respectively), while the right panel presents the comparative results with AVX2-VNNI disabled. First, consistent trends are observed across distinct platforms (as shown in the left and center panels), suggesting that these performance characteristics are intrinsic to the quantization methods rather than artifacts of specific hardware configurations.

Regarding the \q{Q4} series, \q{Q4\_0} consistently outperforms \q{Q4\_k} (ratio $>1$) across all setups, irrespective of VNNI support. This demonstrates its superior computational efficiency, which stems from the streamlined unpacking process inherent to the \q{Q4\_0} implementation~\cite{llama_cpp_2024_q40_q80}. Furthermore, this performance advantage is attenuated during the decoding phase; this reduction aligns with our identification of decoding as a communication-bound task, where computational speedups yield diminishing returns.

Conversely, for the \q{Q5} series, \q{Q5\_k} achieves higher throughput than \q{Q5\_0} when VNNI is enabled. This reversal is attributed to the complex bit-shifting operations required for \q{Q5\_0} unpacking~\cite{llama_cpp_2024_q50_q80}, which offset the parallelization gains typically provided by VNNI. However, this discrepancy becomes less pronounced during decoding, as the dominant communication overhead masks the differences in computational efficiency.

\subsection{System Resource Utilization Results}
\label{sec:results_system_resource_utilization}

\begin{figure}[t]
    \centering
    \begin{subfigure}[t]{0.48\columnwidth}
        \centering
        \includegraphics[width=\linewidth]{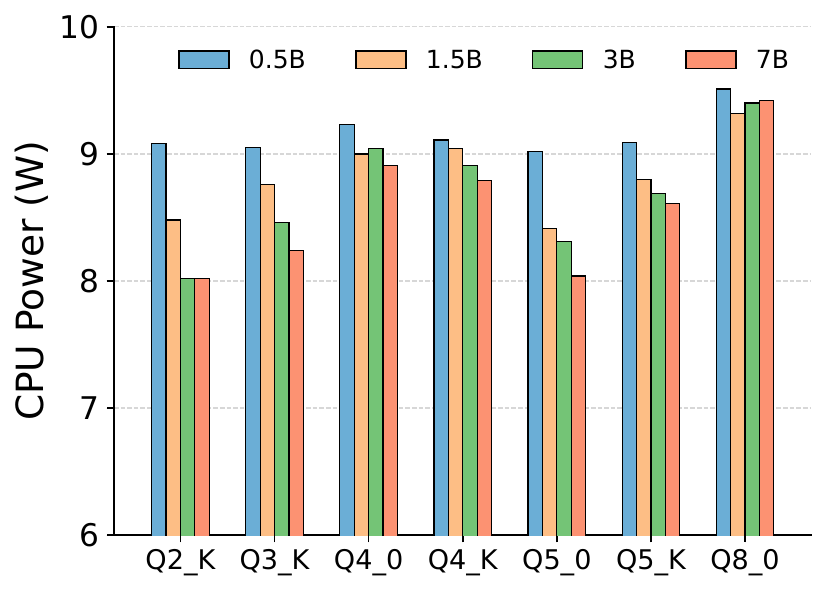}
        \vspace{-5mm}
        \caption{CPU Power Consumption}
        \label{fig:qwen_cpu_power}
    \end{subfigure}
    \hfill
    \begin{subfigure}[t]{0.48\columnwidth}
        \centering
        \includegraphics[width=\linewidth]{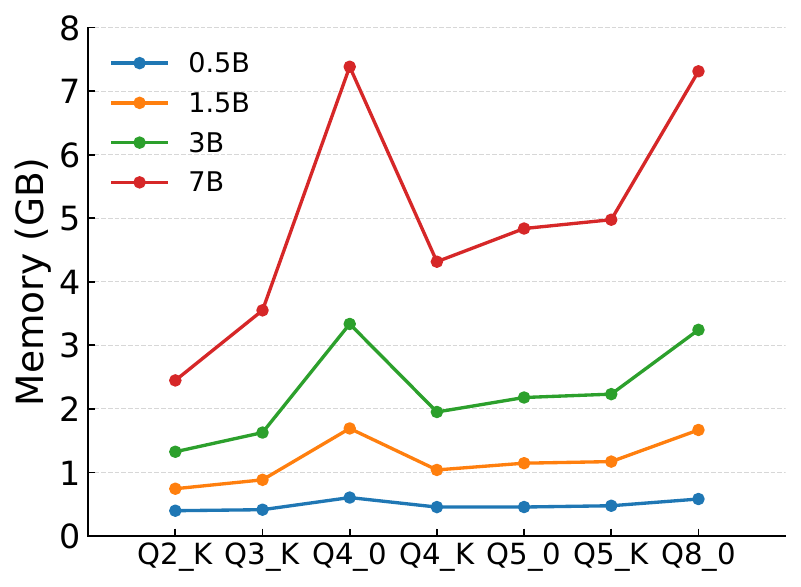}
        \vspace{-5mm}
        \caption{Memory Consumption}
        \label{fig:qwen_memory}
    \end{subfigure}
    \vspace{-2mm}
    \caption{System Resource Utilization for Qwen~2.5 Models with 128-token Inputs and 1000-token Outputs.}
    \label{fig:power_qwen}
\end{figure}

Following the methodology described in \Cref{sec:system_resource_utilization}, we evaluate the CPU power and memory consumption of representative LLMs during continuous execution on a laptop, applying diverse quantization methods. \Cref{fig:power_qwen} illustrates the results for an input length of 128 tokens. Results for the other three input configurations are omitted for brevity, as they exhibit a similar pattern.
We derive three key observations regarding system resource utilization:

1) CPU power consumption correlates with the computational intensity of the quantization methods rather than the BPW.

2) Smaller models with higher computational overhead result in increased CPU power consumption.

3) Memory footprint aligns with the model size and the BPW of the quantization methods.

\textbf{CPU Power Consumption Results}. 
We evaluate the CPU power consumption across various model architectures and quantization methods. 
\Cref{fig:qwen_cpu_power} illustrates the results for different Qwen 2.5 model variants processing 128 input tokens. 
Although power consumption exhibits only minor variations within a narrow range of 7.9\,W--9.5\,W, these fluctuations consistently correlate with the computational complexity of the underlying implementations. 

Analysis of the open-source quantization implementations in \cite{llama_cpp_2024_quantc} reveals that the \q{Q2\_K}, \q{Q3\_K}, and \q{Q5\_K} methods incur higher unpacking overhead compared to \q{Q4\_K} and \q{Q8\_K}. 
For instance, the padding and bit-shift operations required by \q{Q3\_K} increase instruction latency. % Check this sentence
Consequently, computational throughput is reduced, leading to slightly lower CPU power consumption. 
Specifically, on the 1.5B model, \q{Q4\_0} (9.2\,W) and \q{Q8\_0} (9.5\,W) consume marginally more power than \q{Q5\_K} (8.5\,W). 
Further, we monitor the CPU frequency and power consumption for all executions using HWiNFO~\cite{hwinfo64}. These measurements corroborate the patterns observed in \Cref{fig:qwen_cpu_power}. However, we omit the detailed plots due to space constraints.

% We evaluate the computational efficiency of quantized LLMs on edge devices by analyzing CPU power consumption across various bit-widths. As shown in \Cref{fig:qwen-0.5b-64}, power consumption remains relatively stable across all tested quantization levels (\q{Q2\_K} to \q{Q8\_0}), fluctuating within a narrow 7.9W to 9.5W range. 
% However, the \q{Q4\_0} and \q{Q8\_0} configurations exhibit marginally higher power consumption (9.2W and 9.5W, respectively) compared to \q{Q5\_0} (8.5W), suggesting subtle differences in mixed-precision processing efficiency. 
% This discrepancy is likely attributable to hardware-level optimizations for specific arithmetic units. For instance, \q{Q5\_0}'s alignment with 32-bit register boundaries may reduce instruction pipeline stalls, whereas the irregular bit grouping in \q{Q4\_0} could introduce additional cycle overhead for bit-unpacking operations.

% Moreover, quantization BPW fundamentally alters the relationship between model scale and power consumption (\Cref{fig:power_qwen}). 
% At \q{Q2\_K} precision, we observe a paradoxical trend where larger models consume less power; for example, the 7B model uses 8.27W versus 9.33W for the 0.5B model.
% This occurs because the workload shifts from being compute-bound to memory-bound at this low precision.
% Thus, CPU power draw is dictated by the latency of data movement from memory---during which the CPU is often underutilized---rather than by the model's computational complexity.

Moreover, quantization BPW significantly impacts the scaling behavior of power consumption. At \q{Q2\_K} precision, an inverse trend emerges: larger models consume less power (e.g., 7B at 8.27,W vs. 0.5B at 9.33,W). This occurs as the workload shifts from compute-bound to memory-bound. Thus, power consumption is dominated by memory transfer latency and the resulting CPU idle time, effectively decoupling power draw from model size.

% However, the above trend disappears at \q{Q8\_0}, where the 7B and 0.5B models exhibit minimal divergence. We observe two key patterns: 1) Ultra-low bit quantization (\q{Q2\_K}) enables memory access optimization for larger models (e.g., reduced cache misses due to full weight residency), overriding computational load increases; 2) High-efficiency quantization, e.g., \q{Q8\_0}, saturates CPU arithmetic units regardless of model scale, diminishing size-related efficiency variations. The results suggest that aggressive quantization shifts power bottlenecks from computation to memory subsystems.

Conversely, this inverse trend dissipates at \q{Q8\_0}, where the 7B and 0.5B models exhibit negligible divergence in power consumption. 
First, ultra-low bit quantization (e.g., \q{Q2\_K}) facilitates memory access optimizations---specifically reduced cache misses due to lower memory footprints---which outweigh the overhead of increased unpacking logic. 
Second, higher-precision formats such as \q{Q8\_0} saturate CPU arithmetic units regardless of model scale, thereby diminishing size-related variations in efficiency. 
These findings are consistent with our observation in~\Cref{sec:results_deployment_efficiency} that aggressive quantization effectively shifts the system's power bottleneck from the computational units to the memory subsystems.

% Models that are compute-bound consume significantly more power than those that become memory-bound, as the latter leads to reduced CPU utilization and frequency. 
% This principle is evident across various quantization schemes.
% High-power models such as \q{Q4\_K} and \q{Q8\_0} sustain high CPU utilization, particularly during decoding.
% In contrast, low-BPW models like \q{Q2\_K} \q{Q3\_K} become memory-bound, leading to lower power consumption manifested through reduced CPU utilization and increased power. 
% This trade-off also explains specific behaviors: \q{Q4\_0} consumes more power than \q{Q4\_K} due to a compute-intensive prefill phase, while \q{Q5\_0} uses less power than \q{Q5\_K} by inducing a memory-intensive decoding phase. 
% Therefore, we conclude that low-level operational characteristics dictated by the quantization implementation are the primary determinant of CPU power consumption.

\textbf{Memory Consumption Results}. 
Complementing the power analysis, we evaluate the memory requirements of the quantized models. 
As illustrated in \Cref{fig:qwen_memory}, memory consumption generally scales monotonically with the quantization Bit-Per-Weight (BPW). 
For instance, the 0.5B model footprint spans from 392\,MB (\q{Q2\_K}) to 576\,MB (\q{Q8\_0}), while the 7B model ranges from 2411\,MB to 7297\,MB. However, we observe a distinct anomaly regarding the \q{Q4\_0} format, which defies this trend across model scales. 
The 7B \q{Q4\_0} model requires 6910\,MB, a value significantly exceeding the expected interpolation between \q{Q3\_K} (3535\,MB) and \q{Q5\_0} (4837\,MB).

\begin{figure}
    \centering
    \includegraphics[width=1\linewidth]{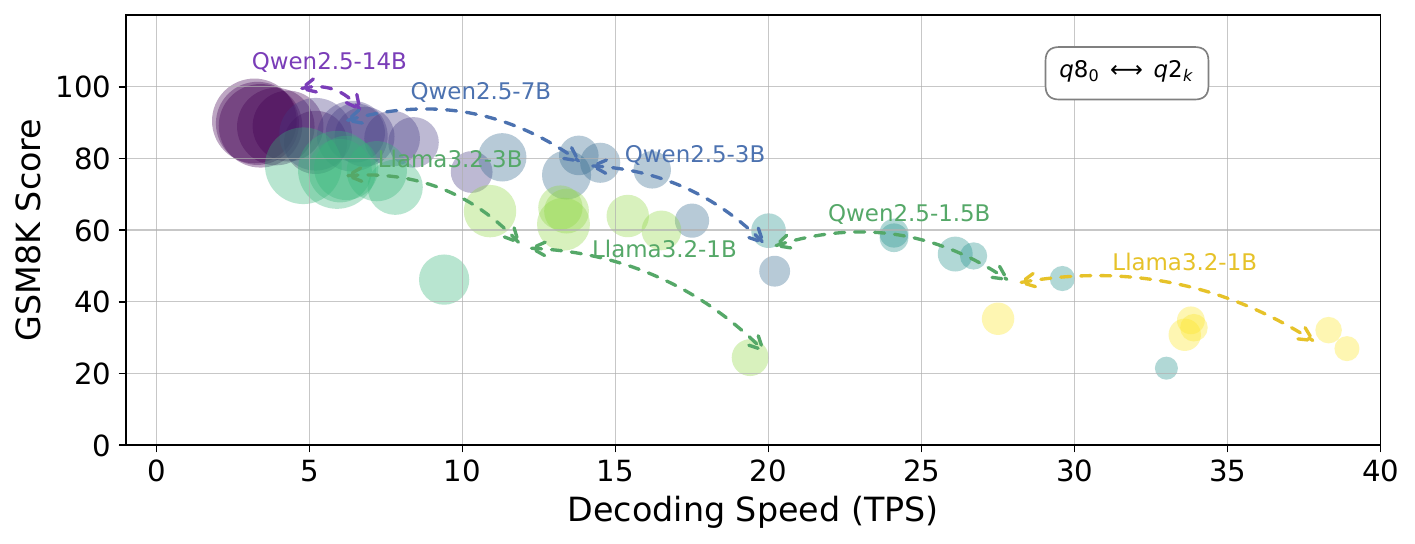}
    \vspace{-6mm}
    \caption{Relationship between Decoding Throughput and Model Capability (From \q{Q8\_0} to \q{Q2\_K}). }
    \label{fig:trade-off}
\end{figure}

\textbf{Pareto Analysis of Performance Metrics.}
We further examine the multi-dimensional trade-offs between accuracy, throughput, and memory consumption, visualized via the Pareto frontier in \Cref{fig:trade-off}. 
In this representation, the marker area corresponds to the on-device memory footprint of each model. The results yield three primary insights. 
First, Qwen 2.5 variants consistently exhibit superior Pareto efficiency compared to Llama baselines across all scales. 
Second, 4-bit quantization emerges as the optimal operating point, accelerating inference throughput by 30--50\% while maintaining near-lossless accuracy. 
Third, pushing compression beyond this threshold (i.e., below 4 bits) results in a precipitous degradation of model performance.

% Moreover, robustness to quantization is scale-dependent. The Qwen2.5 7B model, for instance, maintains accuracy comparable to its 14B counterpart at double the speed, while sub-3B models suffer disproportionate accuracy losses when quantized. 
% Moreover, and most critically, model scale is a more dominant factor than quantization in defining the efficiency-accuracy trade-off. Scaling a model from 7B to 14B parameters sacrifices significant speed for accuracy, while 4-bit quantization on the 7B model provides a substantial speedup with a negligible accuracy drop. This establishes model scaling as the primary determinant of the overall performance balance, with quantization serving as a secondary tool for fine-tuning speed and memory.

\section{Conclusion and Discussion} \label{sec:discussion}
% power analysis
This paper presents a systematic evaluation of LLM inference on edge devices, including model capability, deployment efficiency, and system resource utilization across diverse model sizes and quantization methods on different hardware platforms. 
Our experimental results yield several key findings.

\textit{Model Capability}.
Model capability increases monotonically with Bits Per Weight (BPW) and model architectures. Qwen models generally outperform Llama models. 
Further, large models employing low-bit quantization demonstrate superior performance compared to smaller models utilizing higher-bit quantization.

\textit{Deployment Efficiency}.
LLM inference throughput is primarily governed by an inverse correlation with model size and BPW. 
During the communication-bound decoding phase, higher BPW typically results in a monotonic decrease in throughput. However, this trend is nuanced by the computational cost of the specific quantization method. 
For a fixed BPW, the efficiency of hardware-specific implementation emerges as a critical factor. Moreover, for sufficiently small models, communication bottlenecks can be alleviated, where the reduced model and KV-cache footprints improve cache hit rates and increase throughput. 
During the computation-bound pre-filling phase, the computational efficiency of the selected quantization method becomes paramount for throughput. 
Conversely, for large models during pre-filling, which are generally less computation-bound, effective management of communication overheads is crucial for optimizing throughput.

\textit{System Resource Utilization}.
Memory consumption exhibits a monotonic increase with BPW. 
Quantization methods characterized by fewer CPU operations during the unpacking stage generally correlate with higher overall CPU utilization, as computational throughput is less frequently impeded by this process compared to less efficient alternatives.

\textbf{Insights}.
Due to our results, we formulate guidelines for model and quantization selection, specifically targeting the advancement of on-device framework architectures and efficiency optimization:

\textit{Trade-off between Model Capability and Efficiency}. For large-scale models, low-bit-width quantization typically preserves accuracy while offering only marginal gains in deployment efficiency. 
Conversely, for small-scale models, low-bit-width quantization achieves accuracy comparable to its high-bit-width counterpart while substantially improving deployment efficiency.

\textit{Model Selection for Resource-Constrained Scenarios}.
For applications necessitating high accuracy, we advocate for \textit{large-scale models utilizing moderate quantization} (e.g., 4-bit), which achieve an equilibrium between model fidelity and system overhead. 
In contrast, for latency-sensitive scenarios, \textit{compact models employing similar quantization schemes} prove to be the most effective strategy.

\textit{Bottleneck Identification on Edge Devices}.
Performance constraints on edge devices bifurcate based on model scale. 
Larger models ($>1$B) are predominantly limited by data transfer (communication-bound), making bandwidth optimization critical. 
Smaller models ($<0.5$B) with extended contexts, however, are limited by arithmetic capacity (compute-bound), necessitating strategies focused on computational efficiency.

% \section{Reproducibility Statement} \label{sec:impact}
% This paper is reproducible. The code is open-source. The detailed proofs of theoretical results are available in the Appendix.
% \begin{enumerate}
%     \item 

%     \item 
    
%     \item 
% \end{enumerate}

\appendix
\section{Appendix} \label{sec:appendix}

\subsection{Details of Importance Matrix} 
\label{sec:important_matrix}
To reduce the precision degradation between de-quantized and pre-quantized values, \q{Llama.cpp} formulates an optimization problem that leverages activation statistics to construct weight-aware objective functions. Let $\mathbf{w} \in \mathbb{R}^d$ denote a $d$-dimensional learnable parameter vector, with $\mathbf{q} \in \mathbb{Q}^d$ representing its quantized counterpart obtained through standard quantization procedures. Let $\mathbf{a} \in \mathbb{R}^d$ denotes the corresponding activation vector from the preceding layer. For each dimension $i \in \{1, 2, \dots, d\}$, the per-block quantization problem can be formulated as a quadratic programming problem as $\min_{s,m}\mathbb{E} [ \sum_{i=1}^{d}(\mathbf{q}_i-\mathbf{w}_i)\mathbf{a}_i ]^2$ ($s$ and $m$ are the scale factor and the minimum value defined in \Cref{sec:preliminary}). 
Taking the reformulation in~\citep{intro2024david}, the sum-squared operation in the objective above can be simplified into squared-sum form as $\mathbb{E} [ \sum_{i=1}^{d}(\mathbf{q}_i-\mathbf{w}_i)\mathbf{a}_i ]^2  \approx \mathbb{E} [ \sum_{i=1}^{d}\mathbf{a}_i^2(\mathbf{q}_i-\mathbf{w}_i)^2]$, 
where the coefficient $\mathbf{a}_i$ serves as a weighting factor in the optimization objective. 
\begin{proof}
We expand the quadratic term as $\mathbb{E}\big[ \sum_{i=1}^{d}(\mathbf{q}_i-\mathbf{w}_i)\mathbf{a}_i \big]^2 =
\mathbb{E}\big[
\sum_{i=1}^{d}\big((\mathbf{q}_i-\mathbf{w}_i)\mathbf{a}_i\big)^2
+
\sum_{i=1}^{d}\sum_{\substack{j=1, j\neq i}}^{d}
(\mathbf{q}_i-\mathbf{w}_i)\mathbf{a}_i
(\mathbf{q}_j-\mathbf{w}_j)\mathbf{a}_j
\big].$ 
On the RHS, taking expectation term-by-term yields $\mathbb{E}\big[
\sum_{i=1}^{d}\big((\mathbf{q}_i-\mathbf{w}_i)\mathbf{a}_i\big)^2
\big] + \mathbb{E}\big[
\sum_{\substack{j=1, j\neq i}}^{d}
(\mathbf{q}_i-\mathbf{w}_i)\mathbf{a}_i
(\mathbf{q}_j-\mathbf{w}_j)\mathbf{a}_j
\big]$.
Eliminating the second term yields 
$\mathbb{E}\big[ \sum_{i=1}^{d}(\mathbf{q}_i-\mathbf{w}_i)\mathbf{a}_i \big]^2 \approx
\mathbb{E}\big[
\sum_{i=1}^{d}\mathbf{a}_i^2(\mathbf{q}_i-\mathbf{w}_i)^2
\big]$.
\end{proof}

Empirical studies suggest that parameters with larger magnitudes exhibit greater influence in neural network computations~\citep{imatrix2024jukofyork}. To amplify the significance of these salient parameters, a squared magnitude term $x_i ^2$ is incorporated into the weighting factor~\citep{imatrix2024jukofyork}. $x_i ^2$ also serves as a simple approximation of the diagonal entries of Hessian matrices~\citep{imatrix2024jukofyork}. 
Moreover, to address numerical instability in low-magnitude regimes, block-wise mean squared value of original data is calculated as $\sigma_2 = \frac{1}{n}\sum_{i=1}^n \mathbf{x}_i^2$~\citep{imatrix2024jukofyork}. 
This regularization prevents the systematic underestimation of near-zero parameters during quantization~\citep{imatrix2024jukofyork}. The complete importance matrix is therefore formulated as $\mathbf{\tilde{a}}_i^2 = \mathbf{a}_i^2 \sqrt{\sigma_2  + \mathbf{x}_i^2}$.

%%%%
% Note how the command \texorpdfstring takes effect.
% \texorpdfstring{#1}{#2} will force the compiler to use text in #2 when
% generating bookmarks while display the title in #1 in the text mode.
%
% If the table here '\Cref{tb:...}' has been changed, say, to Table 2, you
% have to manually change parameter #2 to the new one, otherwise the
% pdf bookmark will be incorrect.
%
\subsection{Details of Selected Quantization Methods}
We introduce the \textit{mini-block technique} utilized in \q{Llama.cpp}, which enhances data representation fidelity by quantizing parameters into smaller, independent groups~\cite{llamacppquants2024}. 
LLM parameters are partitioned into contiguous blocks $\mathcal{B}_i$, each of a predefined size $d_1 \in \mathbb{N}^+$. 
Each block $\mathcal{B}_i$ is then quantized independently using its own scale factor $s_i$ and zero-point (denoted as $z_i$). 
Thus, for any parameter $w \in \mathcal{B}_i$, its quantized value $q_w$ is given by $q_w = \mathcal{Q}(w; s_i, z_i)$, where $\mathcal{Q}(\cdot)$ is the quantization function. 
This allows for finer-grained adaptation to the local data distribution within each block.

\label{sec:quants_intro_details}
The \q{Q8\_0}, \q{Q5\_0}, and \q{Q4\_0} methods employ symmetric quantization with a single-layer block structure, grouping 32 weights per block. These methods allocate 8, 5, and 4 bits to weight representation, respectively, while uniformly reserving 16 bits to store quantization parameters (scale and minimum values). Due to the additional bit overhead for shared scale parameters, the BPW for \q{Q8\_0}, \q{Q5\_0}, and \q{Q4\_0} are as 8.5, 5.5, and 4.5 bits, respectively.

The remaining \q{N\_k} quantization methods employ distinct precision-mixed strategies tailored to specific model components, featuring diverse block sizes and bit allocations for quantization parameters. 
% As detailed in~\Cref{tb:llama_cpp_kquant}, 
The \q{Q5\_K} method utilizes symmetric quantization for half of the \q{attention.wv} and \q{feed\_forward.w2} components, structured with a primary block (16 weights) and a secondary block (16 parameters). Here, weights and scale parameters are allocated 6 bits and 8 bits, respectively, resulting in a BPW of 6.5625. For asymmetric quantization, \q{Q5\_K} employs a 32$\times$8 block configuration with 5 bits for weights and 6 bits for parameters, yielding a BPW of 5.5. The component-specific configurations in \q{Q4\_K} mirror those of \q{Q5\_K}, but with reduced bit allocations: 4 bits for weights and 6 bits for scale parameters, achieving a lower BPW of 4.5.

In the \q{Q3\_K} method, components \q{attention.wv}, \q{attention.wo}, and \q{feed\_forward.w2} utilize a 32$\times$8 block configuration, allocating 4 bits for weights and 6 bits for scale parameters, yielding 4.5 BPW. For other components, weights are quantized using a 16$\times$16 block structure with 3 bits for weights and 6 bits for scale parameters, achieving a reduced BPW of 3.4375. 
The \q{Q2\_K} method adopts component-specific configurations analogous to \q{Q4\_K} and \q{Q5\_K}, but with distinct parameterizations. For \q{attention.wv} and \q{feed\_forward.w2}, a 32$\times$8 block is employed with 4 bits for weights and 6 bits for scale parameters. The remaining components leverage a 16$\times$16 block structure, allocating 2 bits for weights and 4 bits for scale parameters, further optimizing the BPW metric.

\begin{table}[tbp]\centering
  \caption{Capability Benchmarks of Selected Llama 3 Models}
  \vspace{-4mm}
  \label{tab:task-llama3-performace}
  \renewcommand{\arraystretch}{0.9}
  \resizebox{\linewidth}{!}{%
  \begin{tabular}{c|l|*{9}{c}}
    \toprule
    \multicolumn{2}{c|}{\multirow{2}{*}{\textbf{Models \& Tasks}}} &\multicolumn{8}{c}{\textbf{Quantization Methods (High BPW $\to$ Low BPW)}} \\
    \multicolumn{2}{c|}{} &\q{fp16} &\q{q8\_0} &\q{q5\_k} &\q{q5\_0} &\q{q4\_k} &\q{q4\_0} &\q{q3\_k} &\q{q2\_k} \\
    \midrule
    \multirow{5}{*}{3.1-8B}
    &GSM8K~\citep{cobbe2021training}        
      &/ &\textbf{77.94} &77.33 &76.57 &76.42 &76.80 &71.95 &46.17 \\
    &HellaSwag~\citep{zellers2019hellaswag} 
      &/ &\textbf{80.49} &80.39 &80.19 &79.85 &79.94 &78.93 &77.37 \\
    &MMLU~\citep{hendrycks2021mmlu}         
      &/ &\textbf{68.42} &68.25 &68.22 &67.57 &66.99 &66.47 &59.46\\
    &HumanEval~\citep{chen2021evaluating}   
      &/ &62.80 &\textbf{63.41} &62.20 &61.59 &61.59 &61.59 &43.29 \\
    &TruthfulQA~\citep{lin2022truthfulqa}   
      &/ &54.40 &54.09 &\textbf{54.42} &53.53 &52.13 &52.78 &45.68 \\
    \midrule
    \multirow{5}{*}{3.2-3B}
    &GSM8K~\citep{cobbe2021training}        
      &64.90 &65.28 &65.28 &\textbf{66.26} &63.91 &61.64 &59.89 &24.41 \\
    &HellaSwag~\citep{zellers2019hellaswag} 
      &\textbf{73.62} &73.55 &73.43 &73.20 &72.60 &72.86 &70.93 &61.29 \\
    &MMLU~\citep{hendrycks2021mmlu}         
      &\textbf{60.83} &60.75 &60.44 &60.30 &60.08 &59.72 &57.09 &46.52 \\
    &HumanEval~\citep{chen2021evaluating}   
      &50.61 &48.78 &50.61 &49.39 &\textbf{51.22} &50.61 &47.56 &26.83 \\
    &TruthfulQA~\citep{lin2022truthfulqa}   
      &51.46 &51.66 &51.82 &50.91 &52.09 &50.26 &\textbf{66.34} &45.85 \\
    \midrule
    \multirow{5}{*}{3.2-1B}
    &GSM8K~\citep{cobbe2021training}        
      &35.03 &\textbf{35.25} &34.80 &32.75 &32.07 &30.78 &26.91 &2.96 \\
    &HellaSwag~\citep{zellers2019hellaswag} 
      &60.94 &\textbf{61.00} &60.76 &60.98 &60.04 &58.94 &58.53 &45.66 \\
    &MMLU~\citep{hendrycks2021mmlu}         
      &46.27 &\textbf{46.35} &45.91 &45.94 &44.29 &44.40 &43.11 &31.35 \\
    &HumanEval~\citep{chen2021evaluating}   
      &34.15 &34.15 &\textbf{35.98} &32.93 &31.10 &29.27 &26.83 &4.88 \\
    &TruthfulQA~\citep{lin2022truthfulqa}   
      &43.39 &43.52 &43.43 &43.47 &\textbf{44.24} &43.51 &40.86 &42.48 \\
    \bottomrule
  \end{tabular}
  }%
\end{table}

\subsection{Details of Benchmark Constructions of Model Capability} \label{se:capability_metric_details}
For GSM8K, there are two types of matching scores: \texttt{flexible-extract} and \texttt{strict-match}. The former retrieves the last number in the response as the final answer to a mathematical question, disregarding the exact pattern of the answer. The latter one strictly matches the first number with a given pattern in the response. We use the maximum number between \texttt{flexible-extract} and \texttt{strict-match} matching scores as the evaluation metric.

For HumanEval, the model's code completion ability is assessed using the \texttt{pass@1} metric. It is given a function signature and a docstring describing the function, then tasked with completing it. The pass rate is determined by counting how many generated code snippets pass all test cases.

HellaSwag, MMLU, and TruthfulQA are all multiple-choice evaluation tasks. We calculate accuracy by evaluating the model's ability to select the correct answer from a set of predefined options. In HellaSwag and MMLU, model capabilities are examined by presenting contextual information without explicit choices. We utilize the logits generated by LLMs to compute the cumulative log probabilities for each candidate response. For TruthfulQA, we adopt the \texttt{mc2} score to measure both truthfulness and informativeness, enabling the model to assign probability values to multiple correct answers.

\subsection{Additional Results: Model Capability} \label{sec:additional_results_model_capability}

We showcase the performance of selected Llama 3 models (1B, 3B, and 8B) across aforementioned tasks GSM8K, HellaSwag, MMLU, HumanEval, and TruthfulQA in~\Cref{tab:task-llama3-performace}. Similar to the observations on Qwen 2.5 models, we conclude that larger models in Llama 3 demonstrate superior performance across all tasks compared to smaller models. For example, on GSM8K, the Llama 3.1 8B \q{fp16} model achieves 78.01\% accuracy, while the Llama 3.2 1B \q{fp16} model reaches only 35.03\%. Similar trends are observed, reflecting the general advantage of model size on capability.

Similarly, within each model size, quantization levels significantly impact performance. Lower quantization levels (e.g. \q{q3\_k}, \q{q2\_k}) lead to noticeable performance declines. For instance, the Llama 3.1 8B model achieves 68.26\% on MMLU with \q{fp16} but drops to 66.47\% with \q{q3\_k} and further plummets to 59.46\% with \q{q2\_k}. Similarly, for the Llama 3.2-1B model, accuracy on MMLU decreases from 46.27\% (\q{fp16}) to 43.11\% (\q{q3\_k}) and sharply to 31.35\% (\q{q2\_k}). The drop in quality at \q{q2\_k} is particularly pronounced, emphasizing the limitations of extreme quantization.

Moreover, Qwen 2.5's 7B models surpass Llama 3 8B models on all five tasks, despite having a comparable number of parameters. Similar to Qwen 2.5, Llama 3 maintains stable performance with moderate quantization levels (e.g., \q{q8\_0}, \q{q5\_k}), but its accuracy degrades significantly under \q{q2\_k}. Qwen 2.5 models exhibit better resilience to quantization changes compared to Llama 3. For example, the Qwen 2.5 14B model on MMLU shows only a minor drop from 79.74\% (\q{q5\_k}) to 78.86\% (\q{q3\_k}), whereas the Llama 3.1 8B model experiences a sharper decline from 68.25\% (\q{q5\_k}) to 66.47\% (\q{q3\_k}). Furthermore, it is noticeable that the quantization \q{q2\_k} of the Llama 3.2 1B model proves to be significantly ineffective for tasks such as GSM8K and HumanEval.

\subsection{Additional Results: Deployment Efficiency}
\label{sec:additional_results_deployment_efficiency}

\begin{figure}[htp]
  \centering
  \includegraphics[width=\linewidth]{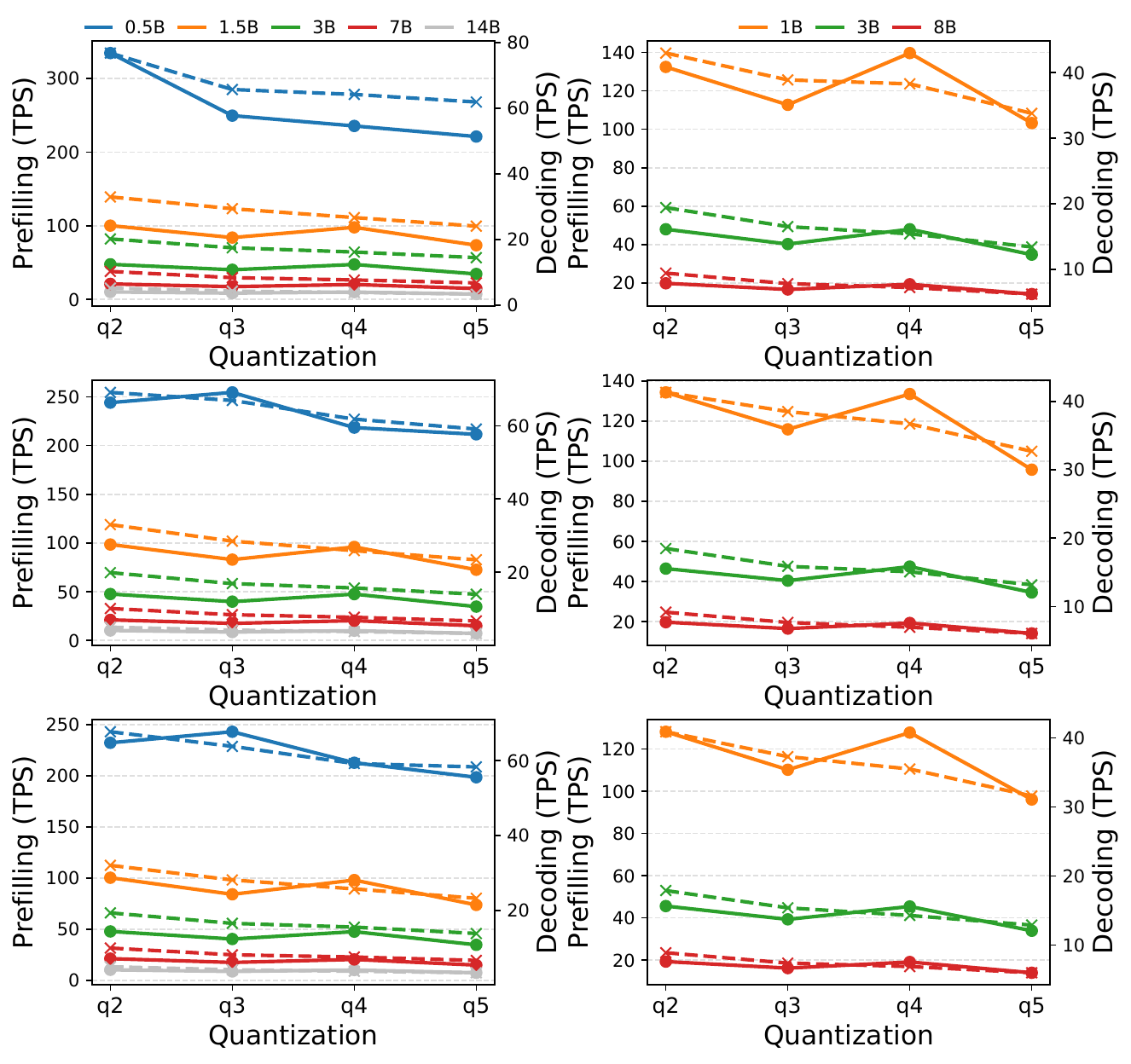}
  \vspace{-6mm}
  \caption{Throughput across Quantization Methods (\q{Q2\_K}, \q{Q3\_K}, and \q{Q4\_K} from Top to Bottom)}
  \label{fig:quan_2_3_4k}
\end{figure}

\Cref{fig:quan_2_3_4k} presents further throughput results of LLMs with different quantization methods, employing selected quantization methods across input token lengths of 64, 256, and 512. 
The results consistently reveal a degradation in decoding throughput with increasing BPW, alongside variations in pre-filling throughput linked to the operational complexity inherent in the quantization methods.

% \input{figs/throughput-tikz/model-throughput-64}
% \input{figs/throughput-tikz/model-throughput-256}
% \input{figs/throughput-tikz/model-throughput-512}

% \begin{figure}[htp]
%   \centering
%   \includegraphics[width=\linewidth]{submitted/figs/throughput_fig/legends/64 token.pdf}
%   \vspace{-0.6cm}
%   \caption{Throughput across Quantization Methods (64-token Inputs)}
%   \label{fig:64_token_throughput}
% \end{figure}

% \vspace{-0.6cm}

% \begin{figure}[htp]
%   \centering
%   \includegraphics[width=\linewidth]{submitted/figs/throughput_fig/legends/256 token.pdf}
%   \vspace{-0.6cm}
%   \caption{Throughput across Quantization Methods (256-token Inputs)}
%   \label{fig:256_token_throughput}
% \end{figure}

% \vspace{-0.6cm}

% \begin{figure}[htp]
%   \centering
%   \includegraphics[width=\linewidth]{submitted/figs/throughput_fig/legends/512 token.pdf}
%   \vspace{-0.6cm}
%   \caption{Throughput across Quantization Methods (512-token Inputs)}
%   \label{fig:512_token_throughput}
% \end{figure}

% combine 3 token graphs

\begin{figure}[htp]
  \centering
  \includegraphics[width=1\linewidth]{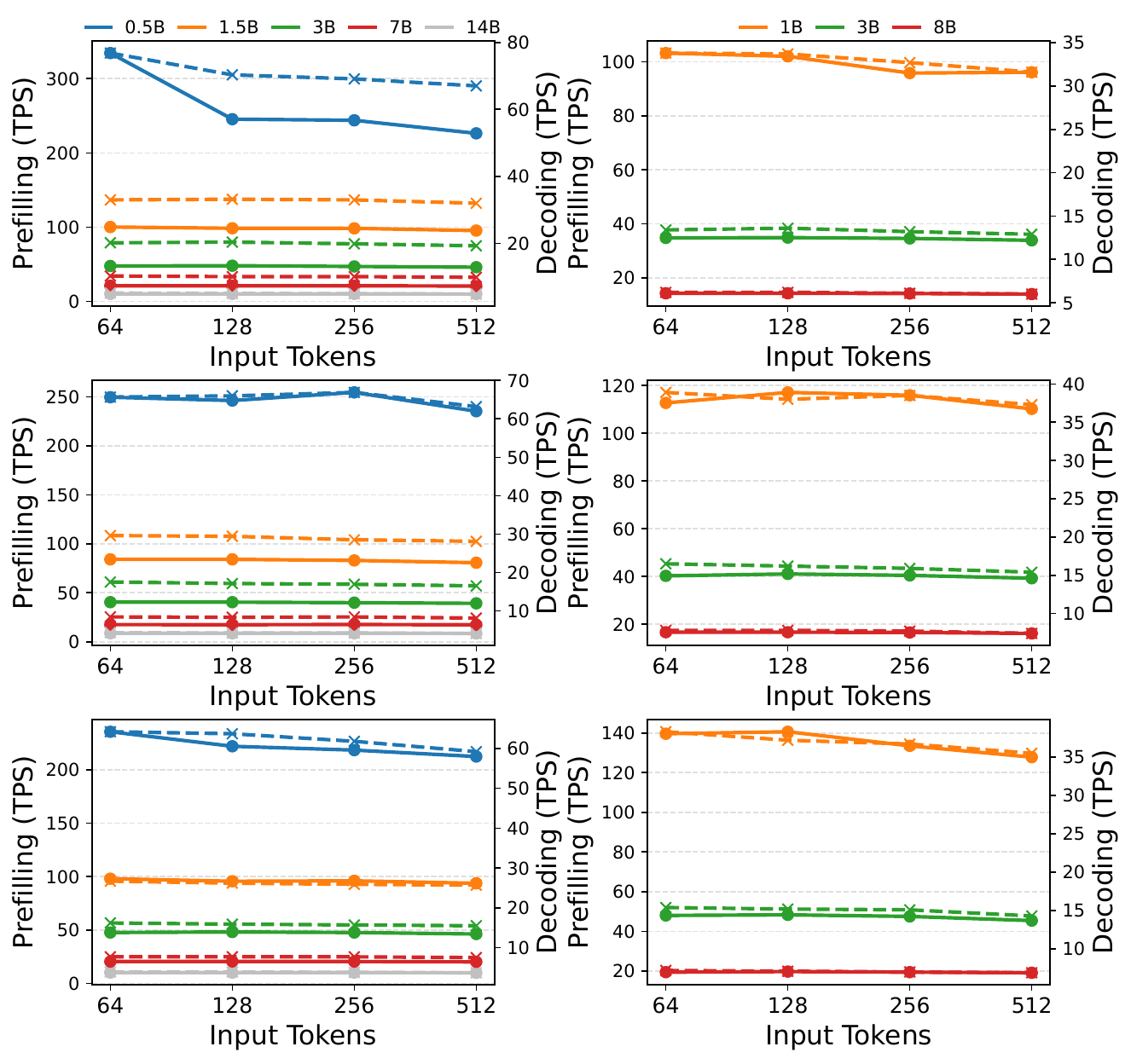}
  \vspace{-6mm}
  \caption{Throughput across Token Lengths (64, 256, and 512-token Inputs from Top to Bottom)}
  \label{fig:token_64_256_512}
\end{figure}

Further, \Cref{fig:token_64_256_512} presents additional results for LLMs across token lengths of 64, 256, and 512, employing quantization at 2-bit, 3-bit, 4-bit, and 8-bit precision levels.
These results consistently reveal an exacerbated trend of throughput degradation with increasing model size, further underscoring the growing impact of computational bottlenecks at larger model scales.
Notably, among 4-bit quantization methods, \q{q4\_0} exhibits superior throughput compared to \q{q4\_k}, an advantage attributed to its implementation involving fewer CPU operations.

\clearpage
\newpage
\bibliographystyle{ACM-Reference-Format}
\bibliography{main}

\end{document}